%% file: main.tex
\documentclass[10pt,twocolumn,letterpaper]{article}

\usepackage[]{cvpr}      

\usepackage{times}  
\usepackage{helvet}  
\usepackage{bbding}
\usepackage{courier}  
\usepackage[hyphens]{url}  
\usepackage{graphicx} 
\usepackage{natbib}  
\usepackage{caption} 
\usepackage{algorithm}
\usepackage{algorithmic}
\usepackage{booktabs}
\usepackage{multirow}
\usepackage{graphicx}

\usepackage{xcolor}
\usepackage{colortbl}
\usepackage{tikz}
\usepackage{enumitem}
\newcommand{\circled}[1]{\tikz[baseline=(char.base)]{
    \node[shape=circle, draw, inner sep=0.5pt,     
          minimum size=0.9em]                      
    (char) {#1};}}

\newlist{circledlist}{enumerate}{1}
\setlist[circledlist]{label=\protect\circled{\arabic*}}
\usepackage{newfloat}
\usepackage{listings}
\usepackage{booktabs}
\usepackage{multirow} 

\usepackage{graphicx}
\usepackage{amsmath,amsthm,amssymb,amsfonts,xspace}

\usepackage{color}

\usepackage{colortbl}

\input{preamble}
\definecolor{cvprblue}{rgb}{0.21,0.49,0.74}
\definecolor{mycolor}{rgb}{0.9,0.9,0.9}
\usepackage[pagebackref,breaklinks,colorlinks,allcolors=cvprblue]{hyperref}

\def\paperID{11448} 
\def\confName{CVPR}
\def\confYear{2026}

\title{LAA3D: A Benchmark of Detecting and Tracking\\ Low-Altitude Aircraft in 3D Space}

\author{Hai Wu$^{1,}$\thanks{Equal contribution.} \quad Shuai Tang$^{2,1,}$\footnotemark[1] \quad Jiale Wang$^{3,}$\thanks{The work done during internship at PCL.} \quad Longkun Zou$^1$ \quad Mingyue Guo$^1$ \\ Rongqin Liang$^1$ \quad Ke Chen$^{1,}$\thanks{Corresponding author.} \quad Yaowei Wang$^1$\\
$^1$Pengcheng Laboratory \quad $^2$South China University of Technology \\ $^3$University of Southern California\\
}
\begin{document}

\makeatletter
\let\@oldmaketitle\@maketitle
\renewcommand{\@maketitle}{
   \@oldmaketitle
	\begin{center}
      \vspace{-6mm}
      \includegraphics[width=1\linewidth]{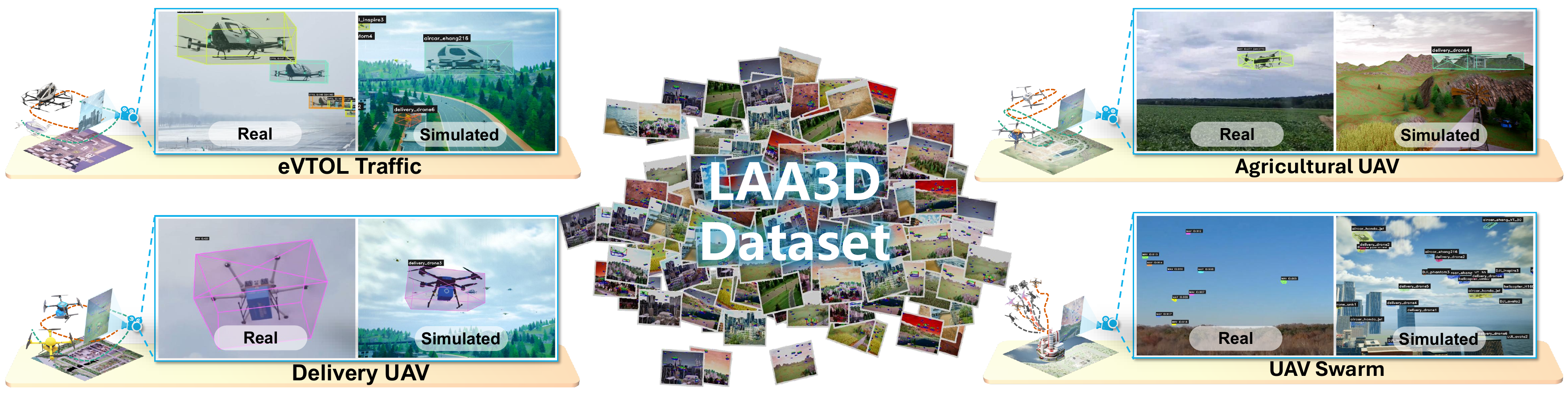}
	\end{center}
   \vspace{-2mm}

  \refstepcounter{figure}\normalfont Figure~\thefigure. 
 LAA3D features a rich variety of real and simulated low-altitude aerial images annotated with 3D bounding boxes, supporting multiple 3D object perception tasks such as 3D object detection and 3D multi-object tracking (MOT).
  \label{fig:abs}
  \newline
  }
\makeatother

\maketitle
\input{sec/0_abstract}    
\input{sec/1_intro}

\input{sec/2_related}

\input{sec/3_method}

\input{sec/4_exp}
\input{sec/5_conclu}
{
    \small
    \bibliographystyle{ieeenat_fullname}
    \bibliography{main}
}
 \input{sec/X_suppl}


\end{document}

%% file: sec/0_abstract.tex
\begin{abstract}
Perception of Low-Altitude Aircraft (LAA) in 3D space enables precise 3D object localization and behavior understanding. However, datasets tailored for 3D LAA perception remain scarce.
To address this gap, we present \textbf{LAA3D}, a large-scale dataset designed to advance 3D detection and tracking of low-altitude aerial vehicles. LAA3D contains 15,000 real images and 600,000 synthetic frames, captured across diverse scenarios, including urban and suburban environments. It covers multiple aerial object categories, including electric Vertical Take-Off and Landing (eVTOL) aircraft, Micro Aerial Vehicles (MAVs), and Helicopters. Each instance is annotated with 3D bounding box, class label, and instance identity, supporting tasks such as 3D object detection, 3D multi-object tracking (MOT), and 6-DoF pose estimation.
Besides, we establish the \textbf{LAA3D Benchmark}, integrating multiple tasks and methods with unified evaluation protocols for comparison. Furthermore, we propose \textbf{MonoLAA}, a monocular 3D detection baseline, achieving robust 3D localization from zoom cameras with varying focal lengths. Models pretrained on synthetic images transfer effectively to real-world data with fine-tuning, demonstrating strong sim-to-real generalization. Our LAA3D provides a comprehensive foundation for future research in low-altitude 3D object perception.
\end{abstract}

%% file: sec/1_intro.tex
\section{Introduction}
In recent years, diverse new aerial vehicles (e.g., delivery drones, electric Vertical Take-Off and Landing (eVTOL), and hybrid-wing aircraft) have emerged, leading to increasingly complex air traffic that challenges traditional regulatory approaches. Conventional rule-based or manual inspection methods are insufficient to meet modern requirements for object recognition, object behavior monitoring, and risk assessment.

Vision-based perception is regarded as one of the core enablers of intelligent low-altitude traffic management~\cite{Artem}. By extracting object categories and motion states from visual data, it supports automated surveillance~\cite{Artem,YeZhengn} of aerial objects. In particular, detecting and tracking objects in 3D space, i.e., estimating object categories, 6-DoF poses, 3D dimensions, and correspondence between frames, enhances situational awareness and potentially improves downstream tasks, such as behavior analysis and  intrusion detection.

\begin{figure}
  \centering
  \includegraphics[width=0.48\textwidth]{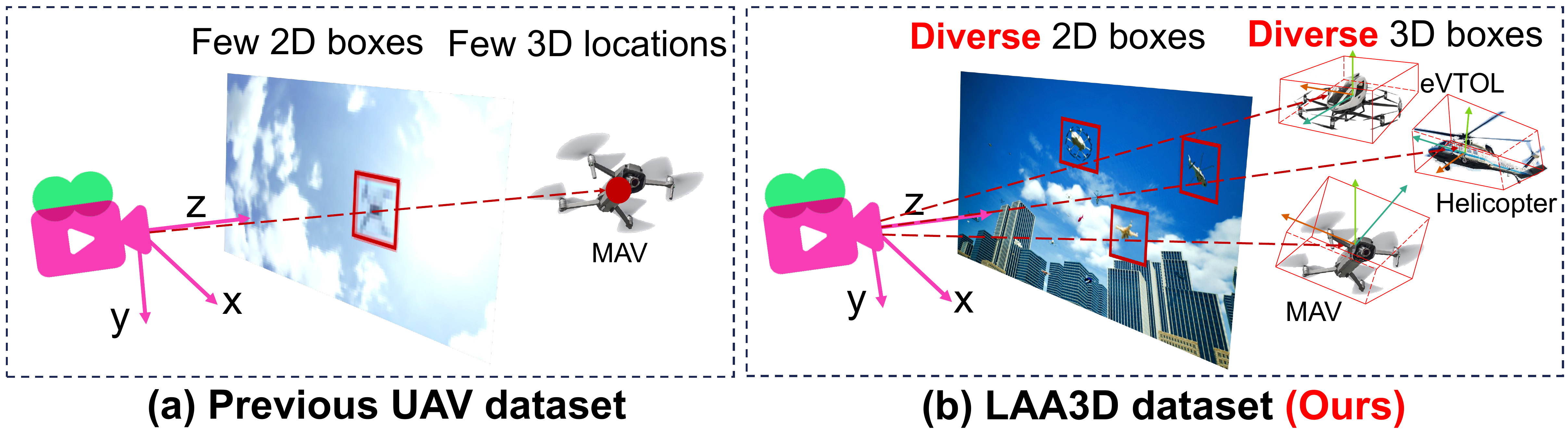}
  \caption{Comparison between LAA3D and previous UAV dataset. (a) The previous UAV~\cite{Mmaud} dataset only contains simple backgrounds and MAV objects. Each scene includes a few objects with limited annotations.  (b) LAA3D contains diverse scenes and various objects, including MAV, eVTOL, and Helicopter. Each scene includes multiple objects with 2D/3D bounding boxes.}
  \label{intro_compare}
\end{figure}

However, building 3D detection and tracking models for low-altitude aircraft (LAA) remains a significant challenge. The primary obstacle lies in the scarcity of datasets that contain diverse LAA instances and 3D annotations. Existing benchmarks, such as Anti-UAV-RGBT~\cite{Anti-UAV-RGBT}, DUT-Anti-UAV~\cite{zhao2022vision}, and Anti-UAV410~\cite{Anti-UAV410}, are restricted to 2D bounding boxes, confining perception to the image plane and limiting 3D understanding. A few datasets include coarse 3D positional data~\cite{Mmaud}, but they typically feature only a single object and lack precise 6-DoF pose supervision (see Fig.~\ref{intro_compare}(a)).  Low-altitude objects often exhibit fast motion and frequent pose variations, making accurate manual labeling of 6-DoF poses extremely challenging.

Moreover, current datasets mainly focus on Micro Aerial Vehicles (MAVs) and fail to capture the growing variety of emerging eVTOLs. Modern platforms such as the EHang 216 for urban air mobility, XPeng’s Voyager X2 with integrated air–ground operations, and SF Express’s hybrid-wing cargo UAV are beginning real-world deployment. Yet, public UAV datasets~\cite{Anti-UAV410,Mmaud,MAV6D} contain no imagery of these next-generation aircraft. This gap stems from the limited availability of real-world data due to the early deployment stage of these vehicles. This data gap impedes progress toward generalized 3D perception models for the rapidly evolving low-altitude airspace.

To overcome these limitations, we propose LAA3D, a large-scale outdoor dataset for low-altitude aircraft perception with high-quality 3D annotations. LAA3D comprises two complementary subsets: LAA3D-real, containing 15,000 real-world images collected from online sources, and LAA3D-sim, containing 600,000 synthetic images generated using the CARLA~\cite{CARLA} simulation platform with a newly developed set of LAA models.
LAA3D covers diverse environments, including urban, coastal, and forest scenes, and a wide range of illumination and weather conditions. Compared with previous UAV datasets~\cite{Mmaud}, LAA3D spans a broader spectrum of aerial object types, including eVTOL, MAVs, and Helicopters (see Fig.~\ref{intro_compare}). It also captures a variety of perception scenarios, such as eVTOL traffic management, delivery drone coordination, agricultural drone monitoring, and drone swarm supervision (see Fig.~\ref{fig:abs}).
For each object, LAA3D provides rich annotations, including 3D bounding boxes with 6-DoF pose, class labels, and instance identities, supporting multiple tasks such as 3D object detection, 3D MOT, and 6-DoF pose estimation.

Beyond dataset construction, we introduce MonoLAA, a monocular 3D detection baseline for images captured by zoom cameras with varying focal lengths, and establish comprehensive 3D object detection and 3D MOT benchmarks. We further evaluate cross-domain generalization from LAA3D-sim to LAA3D-real, demonstrating strong transferability with fine-tuning, and validating the dataset’s effectiveness for real-world applications. We hope that the scale, diversity, and rich annotations of LAA3D will facilitate future advancements in low-altitude 3D object perception. Our contributions are as follows:

\begin{itemize}
\item We present \textbf{LAA3D}, a large-scale outdoor dataset for low-altitude 3D aircraft perception, comprising both real and synthetic frames with precise 3D annotations, providing a comprehensive resource for future research.

\item We establish the \textbf{LAA3D Benchmark}, which integrates state-of-the-art methods under a unified evaluation protocol to enable systematic comparison of 3D LAA detection and tracking performance.

\item We propose \textbf{MonoLAA}, a monocular 3D detection baseline featuring Focal-Length Unification (FLU) and Class-Specific Depth (CSD) designs, achieving robust LAA detection across varying focal lengths.

\item Models pretrained on LAA3D-sim generalize effectively to LAA3D-real with fine-tuning, demonstrating strong sim-to-real transfer and practical deployment potential.
\end{itemize}

\begin{table*}
    \centering
    \setlength{\tabcolsep}{8pt}
    \resizebox{1.\linewidth}{!}{
        

        \begin{tabular}{|l|c|c|c|c|c|c|c|c|}
            \hline
            Dataset & 3D Boxes & 2D Boxes & Indoor/Outdoor & LAA class & Frames & Background & Day\&Weather & Data Type\\ 
            \hline
            Anti-UAV-RGBT~\cite{Anti-UAV-RGBT} & \textcolor{red}{\XSolidBrush} & \textcolor{green}{\Checkmark} & Outdoor & 6 & $\sim$297,000 & 3 & 2 & Real \\
            DUT Anti-UAV~\cite{DUTAntiUAV} & \textcolor{red}{\XSolidBrush} & \textcolor{green}{\Checkmark} & Outdoor & \textbf{35} & $\sim$10,000 & - & 2 & Real \\
            Mittal et al.~\cite{Mittal} & \textcolor{red}{\XSolidBrush} & \textcolor{green}{\Checkmark} & Outdoor & 1 & 2024 & - & 1 & Sim. \\
            Dieter et al.~\cite{Dieter} & \textcolor{red}{\XSolidBrush} & \textcolor{green}{\Checkmark} & Outdoor & 4 & 50,552 & 4 & 1 & Real/Sim. \\
            Anti-UAV410~\cite{Anti-UAV410} & \textcolor{red}{\XSolidBrush} & \textcolor{green}{\Checkmark} & Outdoor & 1 & 150,000 & 6 & 4 & Real \\
            MMAUD~\cite{Mmaud} & \textcolor{red}{\XSolidBrush} & \textcolor{green}{\Checkmark} & Outdoor & 5 & $\sim$51,000 & 1 & 1 & Real \\
            M3D~\cite{M3D} & \textcolor{red}{\XSolidBrush} & \textcolor{green}{\Checkmark} & Outdoor & 10 & 83,999 & 8 & 1 & Real/Sim. \\
            DrIFT~\cite{DrIFT} & \textcolor{red}{\XSolidBrush} & \textcolor{green}{\Checkmark} & Outdoor & 7 & 47,991 & 3 & 4 & Real/Sim. \\
            MAV6D~\cite{MAV6D} & \textcolor{green}{\Checkmark} & \textcolor{red}{\XSolidBrush} & Indoor & 2 & 33,489 & 1 & 1 & Real \\
            \hline
            LAA3D (Ours) & \textcolor{green}{\Checkmark} & \textcolor{green}{\Checkmark} & Outdoor & 26 & \textbf{615,000} & \textbf{8} & \textbf{8} & \textbf{Real/Sim.} \\
            \hline
        \end{tabular}

    }
    \caption{Comparison with recent published low-altitude aircraft  datasets. }
    \label{dataset_table}
\end{table*}

%% file: sec/2_related.tex
\section{Related Work}
\textbf{Anti-UAV datasets.}
The rise of Unmanned Aerial Vehicle (UAV) threats has driven the creation of specialized Anti-UAV datasets~\cite{MAVVID,HalmstadDrone,Anti-UAV600}. Key benchmarks include Anti-UAV-RGBT~\cite{Anti-UAV-RGBT} for cross-modal tracking, DUT-Anti-UAV~\cite{zhao2022vision} for small object detection, the large thermal infrared Anti-UAV410~\cite{Anti-UAV410} tackling tiny objects and clutter, the M3D~\cite{M3D} integrates real-world noise modeling for domain-adaptive MAV detection, and MMAUD~\cite{Mmaud} with multi-sensor integration for realism. The annotations of most datasets are typically restricted to 2D boxes. While MMAUD offers 3D position, it lacks object orientation and typically contains only a few objects per frame, restricting multi-object 3D detection research. In contrast, our LAA3D dataset provides abundant objects with full 6-DoF poses, enabling more comprehensive studies.

\textbf{Monocular 3D object detection datasets.} 
Monocular 3D object detection has progressed with the support of large-scale datasets such as KITTI~\cite{kitti}, nuScenes~\cite{nuscenes}, Argoverse~\cite{Argoverse}, ApolloScape~\cite{apolloscape},  ONCE~\cite{ONCE}, and Waymo~\cite{waymo}, which provide 3D bounding boxes primarily in the road plane, limited to yaw orientation. However, the absence of full 6-DoF poses, especially roll and pitch, restricts fine-grained 3D understanding. Moreover, these datasets lack coverage of LAA objects like drones and eVTOLs. To bridge these gaps, our LAA3D dataset introduces rich 6-DoF annotations for aerial vehicles, offering a new benchmark for monocular 3D perception of air objects.

\textbf{6-DoF pose estimation datasets.} 
Existing 6-DoF object pose estimation datasets~\cite{Bop,Objectron,DexYCB} predominantly focus on static, ground-based objects, offering rich annotations and object diversity. Representative examples include LineMOD~\cite{LineMOD} and YCB-Video~\cite{YCB-Video}, which provide RGB-D sequences and precise pose labels. Other datasets, such as T-LESS~\cite{T-LESS}, address challenges for low texture objects under industrial settings, and ObjectNet3D~\cite{ObjectNet3D} provides category-level 3D pose annotations using real images from diverse indoor and outdoor scenes. However, none of these datasets address challenges posed by aerial objects operating at low altitudes.
The most relevant public dataset to our work is MAV6D~\cite{MAV6D}, which provides pose annotations for micro aerial vehicles (MAVs). Nonetheless, MAV6D is limited to indoor scenes and suffers from a small amount of data. Unlike that, LAA3D provides large-scale outdoor data with 3D annotation, potentially facilitating the research of outdoor LAA 3D detection and tracking.

%% file: sec/3_method.tex
\section{The LAA3D Dataset}
To facilitate research on LAA perception in 3D space, we develop the LAA3D dataset. LAA3D consists of LAA3D-real and LAA3D-sim. The numbers of frames and object classes are more diverse than recent published datasets (see Table~\ref{dataset_table}).  We detail the dataset as follows.

\subsection{LAA3D-real}

\textbf{Data collection.}
We first collected diverse real-world video sequences featuring newly emerging aircraft types (e.g., EHang 216) and various backgrounds from publicly available online sources. Most of these videos consisted of concatenated short clips captured from different viewpoints. To ensure geometric consistency and support tasks such as multi-frame tracking, we decomposed each video into sub-sequences, with each sub-sequence representing a single coherent scanning scene.
For accurate geometric calibration, we estimated camera intrinsics and extrinsics using VGGT~\cite{VGGT}. For sequences with clear backgrounds, we further refined the camera parameters through VGGT+BA~\cite{VGGT}. In total, 185 sequences comprising 15,000 high-resolution frames (1920×1280) were processed.

\begin{figure}
  \centering
  \includegraphics[width=0.48\textwidth]{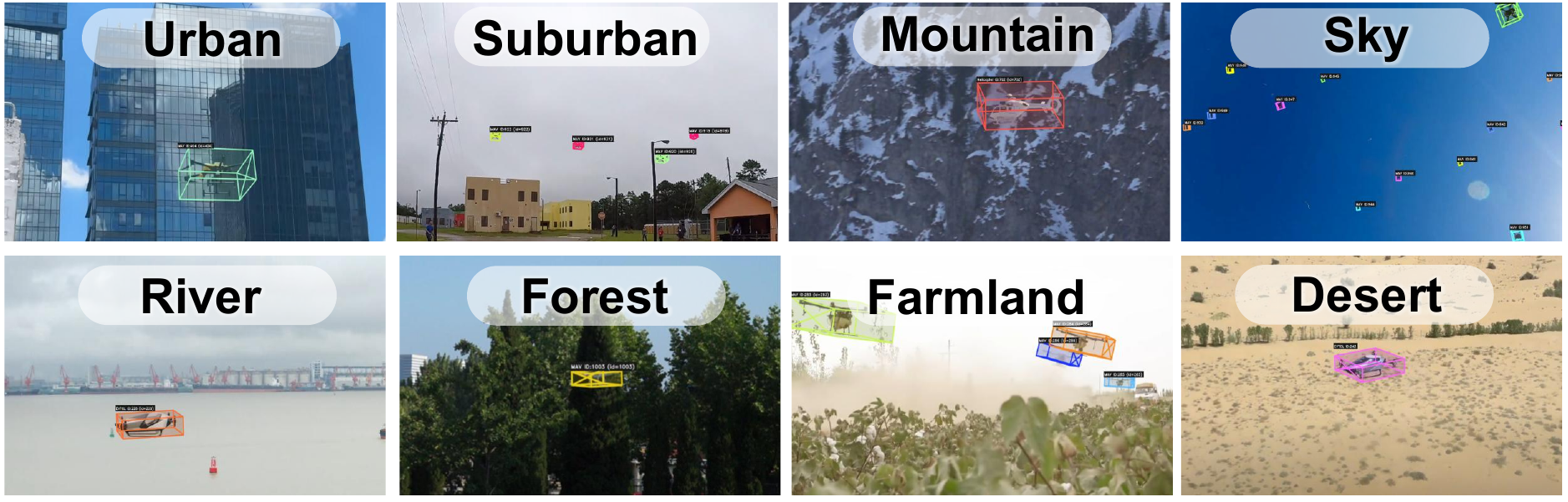}
  \caption{Representative samples from LAA3D-real, covering diverse real-world backgrounds.  }
  \label{real_example}
\end{figure}
\begin{figure}
  \centering
  \includegraphics[width=0.48\textwidth]{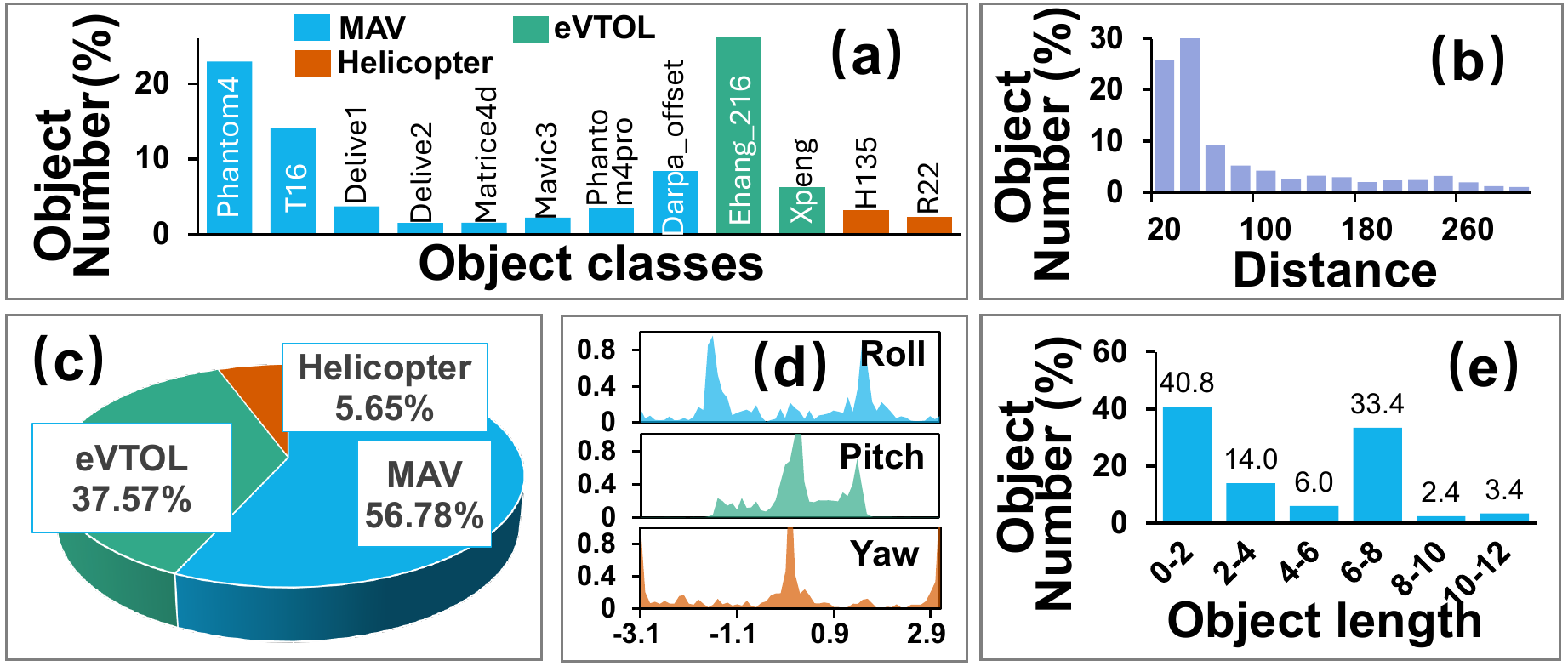}
  \caption{Statistics of LAA3D-real: (a) fine classes distribution, (b) distance distribution, (c) coarse category distribution, (d) orientation angle distribution, and (e) object length distribution.}
  \label{real_statistics}
\end{figure}

\textbf{Data labeling.}
Monocular images lack depth or LiDAR references, making 3D labeling highly challenging. To overcome this, we developed a keypoint-based 3D annotation system (see Supp. Material). Specifically,
for each aircraft type, such as EHang 216, we constructed a keypoint-based CAD model reflecting its true dimensions and geometry. Professional annotators manually aligned these models to real images by adjusting 6-DoF poses until both the 3D bounding boxes and keypoints precisely matched the object’s projected geometry and contours in the image.
Each object is annotated with its 6-DoF pose, 3D dimensions, unique ID, and category label, providing a comprehensive source for 3D detection, MOT, and pose estimation tasks.

\textbf{Object statistics and data split.}
LAA3D-real employs three coarse categories (MAV, eVTOL, and Helicopter), consisting of multiple fine-grained aircraft types (see Fig.~\ref{real_statistics}(a)(c)).
Objects are distributed within 0–300m from the camera, with most appearing within 100 m (see Fig.~\ref{real_statistics}(b)).
Object sizes range from 0–12m, where objects (0–2m) account for around 40\% of all instances (see Fig.~\ref{real_statistics}(e)).
The dataset is divided into training (60\%), validation (20\%), and test (20\%) sets, maintaining consistent background and category distributions across splits (see Supp. Material).

\textbf{Diversity of scenes.}
LAA3D-real spans a wide variety of real-world backgrounds, such as mountainous regions, farmlands, deserts, and skylines (see Fig.~\ref{real_example}).
It also encompasses diverse perception scenarios, such as eVTOL traffic monitoring, drone delivery coordination, agricultural drone inspection, and multi-drone swarm supervision, capturing rich visual, geometric, and environmental variability.
This diversity ensures that LAA3D-real provides a challenging benchmark for assessing model robustness, generalization, and transferability from simulation to the real world.

\begin{figure}
  \centering
  \includegraphics[width=0.48\textwidth]{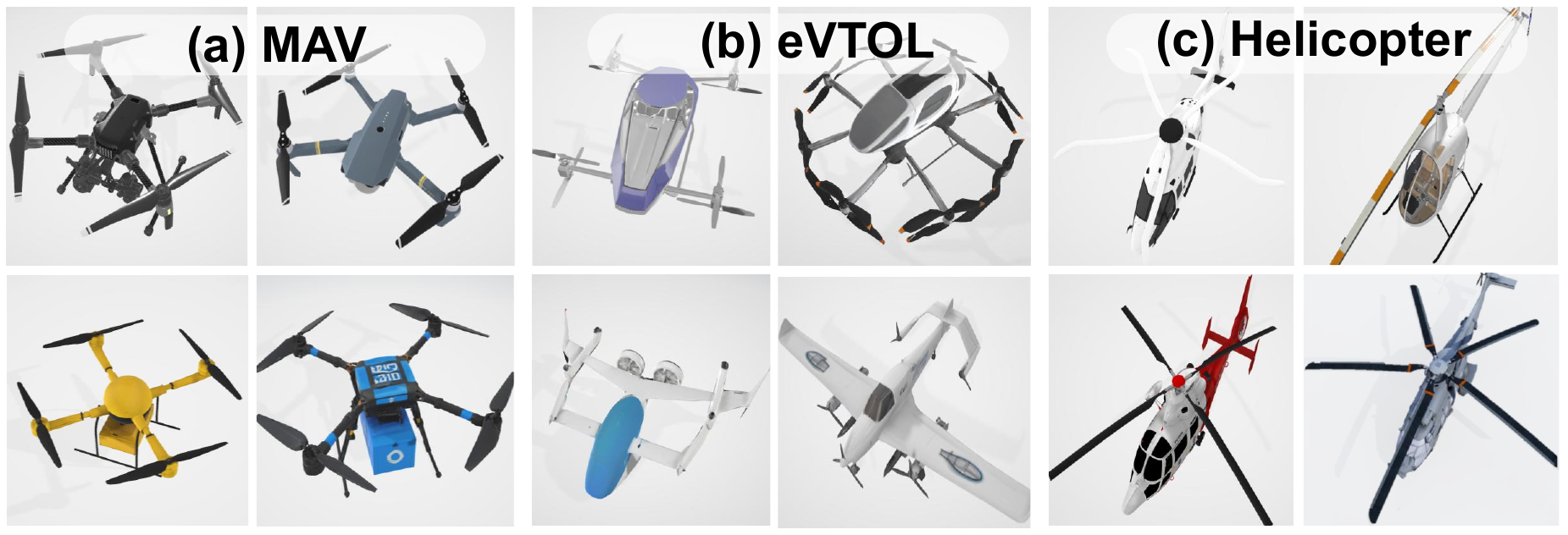}
  \caption{Examples of LAA model set. (a) MAV models consist of consumer drones (e.g, DJI Mavic) and industrial drones (e.g., delivery drone and agricultural drone). (b) eVTOL models include quadrotors (e.g., EHang 216) and hybrid wing aircraft (e.g., EHang VT-30). (c) Helicopter models include consumer aircraft such as H135, R22, etc. }
  \label{sim_uav}
\end{figure}

\subsection{LAA3D-sim}

Despite the construction of a real-world dataset, its scale and background diversity remain limited, while expanding real data collection incurs prohibitive costs. To address this challenge, we introduce LAA3D-sim, a large-scale photorealistic synthetic dataset designed for model development, validation, and pretraining.

\textbf{LAA model library.}
While several prior studies have introduced aircraft simulation systems, the available open-source 3D assets cover only a narrow range of aircraft and lack emerging categories such as diverse eVTOLs. To overcome this, we built a dedicated 3D model library for low-altitude aircraft perception. Professional modelers reconstructed high-fidelity meshes based on real-world geometry and dimensions. In total, 26 aircraft models were created, including 16 MAVs (e.g., DJI Mavic, Inspire 3), 4 eVTOLs (e.g., EHang 216, VT30), and 6 Helicopters (e.g., H135, H160), as shown in Fig.~\ref{sim_uav}.

\begin{figure}
  \centering
  \includegraphics[width=0.48\textwidth]{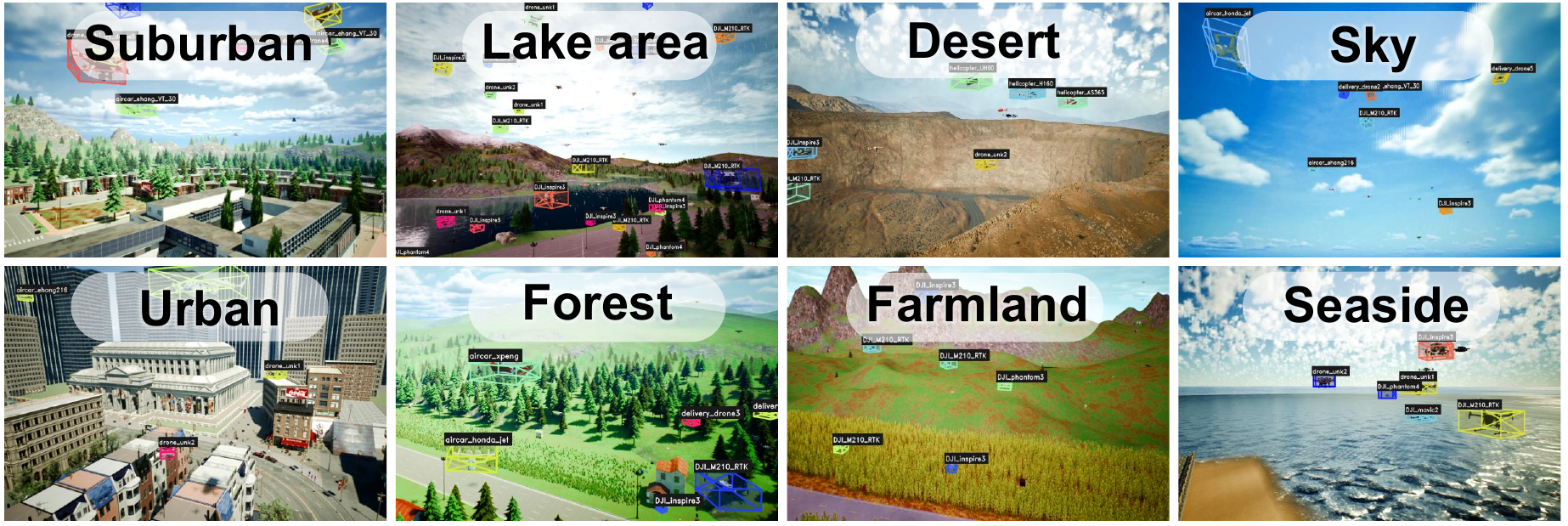}
  \caption{Representative samples from LAA3D-sim covering diverse simulated backgrounds.}
  \label{sim_example}
\end{figure}
\begin{figure}
  \centering
  \includegraphics[width=0.48\textwidth]{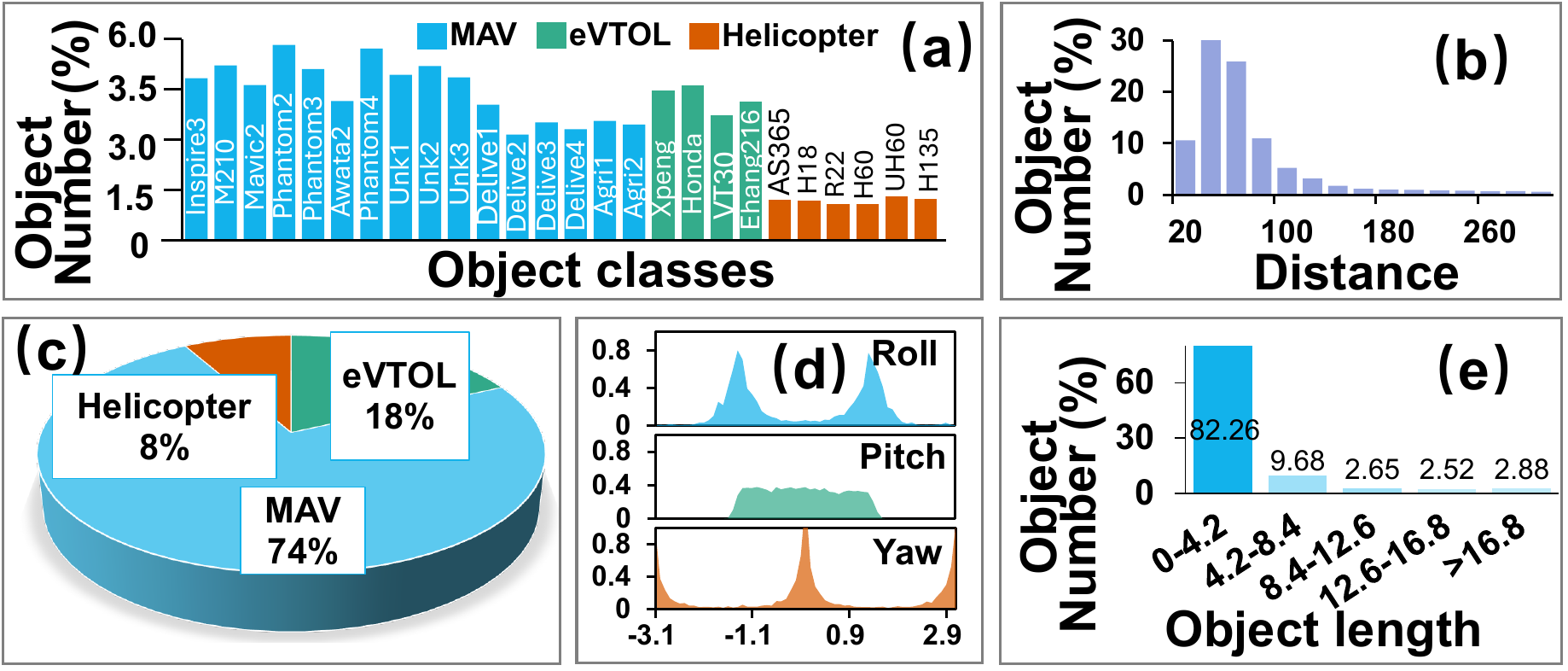}
  \caption{Statistics of LAA3D-sim: (a) fine classes distribution, (b) distance distribution, (c) coarse category distribution, (d) orientation angle distribution, and (e) object length distribution.}
  \label{sim_statistics}
\end{figure}

\textbf{Simulation platform.}
We adopt CARLA~\cite{CARLA} as the base simulator for LAA3D-sim due to its rich, photorealistic environmental assets. Since CARLA natively targets ground vehicles, we extended it by developing custom aerial agents with realistic dynamics and control APIs. These agents perform controlled flights across diverse urban and suburban scenes.
To mimic realistic surveillance settings, we deployed virtual monitoring cameras, both static (on rooftops and towers) and dynamic (mounted on moving platforms), capturing wide-FOV, multi-view image sequences. This configuration provides diverse and representative data for training robust aerial perception models.

\textbf{Data recording and annotation.}
We also simulate diverse perception scenarios with various weather and illumination conditions.
An integrated annotation interface exports synchronized aircraft states and sensor parameters in real time. The dataset comprises $\sim$600k images (1280$\times$720) across 267 sequences at 10 FPS, containing 12.35M annotated objects ($\sim$20 per frame). Each object is labeled with 6-DoF pose, 3D size, category, and instance ID, along with camera parameters and derived 2D boxes.

\textbf{Object statistics and data split.}
To align structurally with LAA3D-real, LAA3D-sim adopts three coarse categories (see Fig.~\ref{sim_statistics}(a)(c)), enabling direct cross-domain evaluation and synthetic-to-real transfer.
Objects are spatially distributed within 0–300m of the camera, with most concentrated within 100~m, closely matching real aerial surveillance ranges (see Fig.~\ref{sim_statistics}(b)). The dataset is dominated by MAVs (see Fig.~\ref{sim_statistics}(e)) and exhibits rotation patterns consistent with real-world distributions (see Fig.~\ref{real_example}(d) and Fig.~\ref{sim_statistics}(d)).
For fair benchmarking, data are split into training (60\%), validation (20\%), and test (20\%) sets, preserving balanced scene and category statistics (see Supp. Material).

\textbf{Diversity of scenes.}
To ensure realism and generalization, LAA3D-sim spans diverse environments, such as coastal, forest, and farmland areas (see Fig.~\ref{sim_example}).
Leveraging CARLA’s dynamic sky engine, LAA3D-sim further introduces extensive weather and illumination variations, clear, overcast, and rainy conditions from sunrise to nighttime, ensuring models trained on it exhibit robust and transferable performance in complex low-altitude environments.

\begin{figure}
  \centering
  \includegraphics[width=0.48\textwidth]{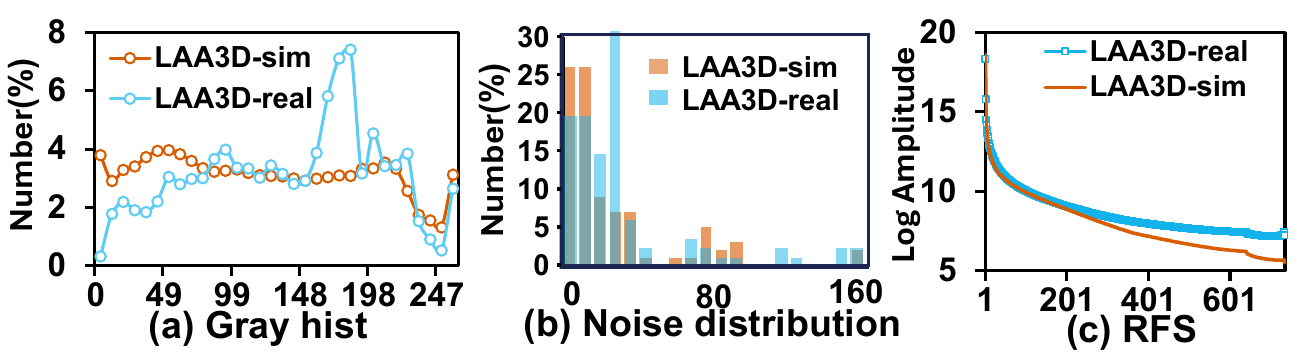}
  \caption{(a–c) Comparisons of grayscale histograms, noise distributions, and radial frequency spectra between LAA3D-real and LAA3D-sim.}
  \label{domain_statistics}
\end{figure}
\subsection{Domain Gap Analytic}
To facilitate future research, we perform a domain gap analysis between LAA3D-sim and LAA3D-real from three perspectives: image-level appearance, object-level characteristics, and scene distribution. 

\textbf{Image-level differences.}
As shown in Fig.~\ref{domain_statistics}, the gray histogram in (a) reveals a sharper intensity concentration in LAA3D-real, indicating richer contrast variations compared to synthetic textures in LAA3D-sim. The noise distribution in (b) shows real images contain stronger and more diverse sensor noise patterns, while simulated data exhibit cleaner pixel statistics due to rendering idealization. The radial frequency spectrum (RFS) in (c) shows LAA3D-real maintains higher high-frequency amplitudes, suggesting sharper edges and more fine-grained textures, whereas LAA3D-sim tends to be spectrally smoother.


\textbf{Object and scene distribution.}
Both datasets share the same taxonomic structure (MAV, eVTOL, and Helicopter) and exhibit similar object–camera distance distributions, with most instances appearing within 100m. Their orientation distributions also align closely. Regarding object size, LAA3D-sim covers a broader range, extending up to 19m, encompassing the 12m range of LAA3D-real. From a scene perspective, LAA3D-sim mirrors the environmental diversity of LAA3D-real, spanning farmland, mountainous, and urban landscapes.

%% file: sec/4_exp.tex
\section{LAA3D Applications}
LAA3D offers comprehensive annotations, supporting tasks such as 3D detection, tracking, and 6-DoF pose estimation.  
This main paper focuses on detection and tracking. Baselines for other tasks are available in the Supp. Material.

\begin{figure}
  \centering
  \includegraphics[width=0.48\textwidth]{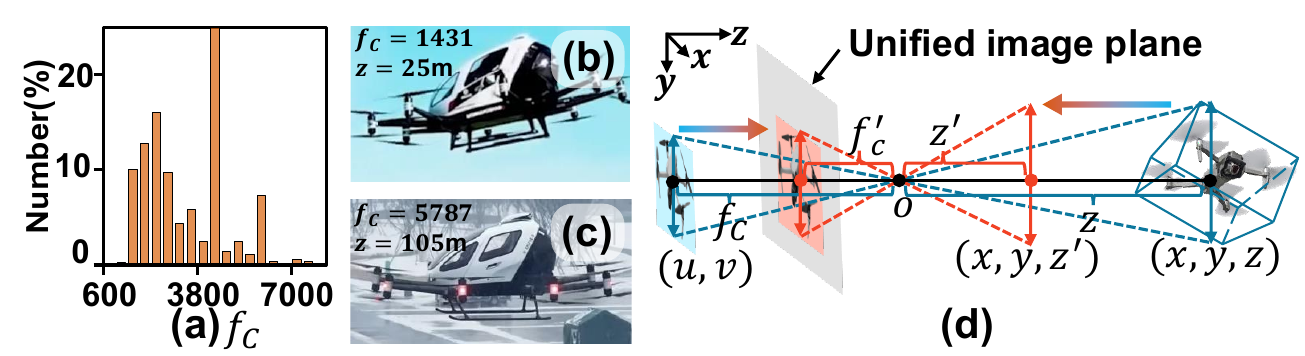}
  \caption{(a) Focal
length distribution of LAA3D-real dataset. (b-c) Objects with similar image-plane sizes but varying depths. (d) Images aligned to a unified plane with the same focal length. }
  \label{method_motivation}
\end{figure}

\subsection{3D Object Detection}

\textbf{Aircraft Detection Score (ADS) metric.} 
Conventional 3D object detection metrics, such as 3D AP~\cite{kitti}, rely on 3D IoU associations between predictions and ground truth, which are less effective for small objects without IoU. 
Inspired by the NDS metric~\cite{nuscenes}, we define our ADS metric based on distance associations, with several adaptations for our dataset. 
(1) Larger class-specific association thresholds.
Given the large variation in object size and detection distance, we employ class-dependent thresholds. 
Since the evaluation distances for MAV, eVTOL, and Helicopter are approximately 2$\times$, 3$\times$, and 6$\times$ greater than those in nuScenes, we set the distance thresholds to [1, 2, 4, 8], [1.5, 3, 6, 12], and [3, 6, 12, 24] meters, respectively.
(2) Incorporating detection recall.
The original NDS combines mAP with true-positive metrics but places limited emphasis on missed detections. 
As small objects in our dataset are prone to being missed, we integrate the detection recall rate into the ADS to provide a more balanced evaluation. 
The final ADS is defined as
$ADS = \frac{1}{8}\left(4mAP + 100\sum_{mTP}(1 - \mathcal{N}(mTP)) + mDR\right)$,
where $mAP$(\%) is the average AP across all association thresholds and classes, $mTP$ represents the average rotation, translation, and size errors of true positives, and $mDR$(\%) denotes the average detection recall. 
$\mathcal{N}(\cdot)$ is the min-max normalization function, with maximum rotation, translation, and size errors of [4, 6, 12]m, [45, 45, 45]$^\circ$, and [0.5, 0.5, 0.5]m for [MAV, eVTOL, Helicopter], respectively.

\subsubsection{MonoLAA Baseline} 
Existing monocular 3D detectors~\cite{Monocd,MonoMAE,MonoRCNN,MonoFlex,MonoCon} target autonomous driving with fixed focal lengths and relatively small detection range. In contrast, low-altitude imaging involves large focal variations (see Fig.~\ref{method_motivation} (a)), diverse object shapes and ranges. To tackle this, we propose MonoLAA, a effective baseline with two key designs: Focal-Length Unification (FLU) and Class-Specific Depth (CSD) modules.

\begin{figure}
  \centering
  \includegraphics[width=.49\textwidth]{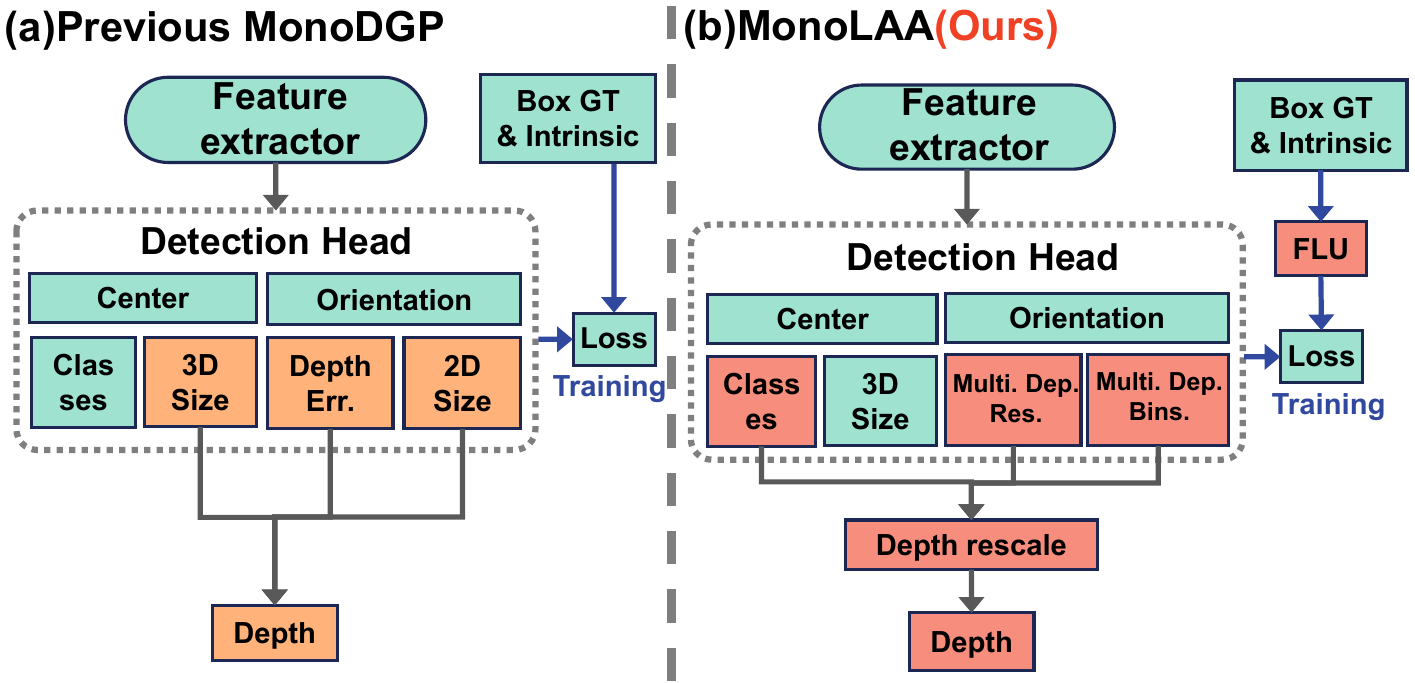}
  \caption{(a) Previous MonoDGP directly regresses the depth components. (b) Our MonoLAA predicts the class-specific and scaled depth after Focal Length Unification (FLU).}
  \label{det_baseline2}
\end{figure}

\textbf{Focal length unification.} 
Near-range objects captured with short-focus lenses can appear the same size in the image plane as distant objects captured with long-focus lenses, making direct depth regression from variable-focal images challenging (Fig.~\ref{method_motivation} (b-c)). A solution is to directly feed the focal length $f_c$ into the detector~\cite{3D-MOOD}. However, due to the typically large variation of $f_c$ in LAA observation and potential unseen $f_c$ ranges at test time, depth predictions remain unreliable. To address this, we propose Focal-Length Unification (FLU): during training, object depths $z$ are scaled to a unified focal length $f_c'$ via $z' = \frac{f_c' z}{f_c}$ (see Fig.~\ref{method_motivation}(d)), and at test time, the predicted depth is rescaled as $z = \frac{f_c z'}{f_c'}$. Here, we adopt $f_c' = 640$, corresponding to a 1280×720 image with 90° FOV.

\textbf{Class-specific depth estimation.} 
Previous 3D detectors use a unified depth encoder for all object categories (Fig.~\ref{det_baseline2}(a)). In low-altitude scenarios, MAVs appear at much shorter ranges than eVTOLs or Helicopters, making unified normalization produce small depth targets and reducing their contribution to the training loss. To address this, we propose a Class-Specific Depth (CSD) head (Fig.~\ref{det_baseline2}(b)), assigning prediction ranges per class: 100m for MAVs, 150m for eVTOLs, and 300m for Helicopters, with each head predicting depth bins plus residuals.

\begin{table}[htbp]
\centering
\small
\setlength{\tabcolsep}{4pt}
\resizebox{\linewidth}{!}{
\begin{tabular}{|l|l|>{\columncolor{blue!10}}c|c|c c c|}
\hline
Backbone & Method & \textcolor{red}{\textbf{ADS}} & mAP & MAV & eVTOL & Heli. \\
\hline
\multirow{5}{*}{ViT-L} 
& CenterNet~\cite{CenterNet} & 46.53 & 32.18 & 35.20 & 36.70 & 24.65 \\
& Center-DORN~\cite{DORN} & 42.66 & 35.33 & 38.13 & 46.17 & 21.50 \\
& MonoDETR~\cite{MonoDETR} & 47.98 & 34.72 & 49.43 & 21.02 & 33.71 \\
& MonoDGP~\cite{MonoDGP} & 48.58 & 30.67 & \textbf{52.03} & 27.59 & 12.40 \\
& MonoLAA(Ours) & \textbf{51.46} & \textbf{41.95} & 45.16 & \textbf{46.25} & \textbf{34.44} \\
\hline
\multirow{5}{*}{ResNet152} 
& CenterNet~\cite{CenterNet} & 49.65 & 36.65 & 43.09 & 37.63 & 29.24 \\
& Center-DORN~\cite{DORN} & 49.98 & 40.83 & 47.19 & 44.75 & 30.56 \\
& MonoDETR~\cite{MonoDETR} & 43.86 & 29.20 & 49.02 & 22.57 & 16.02 \\
& MonoDGP~\cite{MonoDGP} & 50.71 & 35.18 & 47.08 & 31.94 & 26.53 \\
& MonoLAA(Ours) & \textbf{55.23} & \textbf{43.77} & \textbf{49.43} & \textbf{47.53} & \textbf{34.34} \\
\hline
\multirow{5}{*}{ConvNeXt-L} 
& CenterNet~\cite{CenterNet} & 54.48 & 40.20 & 51.14 & 39.28 & 30.18 \\
& Center-DORN~\cite{DORN} & 54.57 & 44.90 & 54.54 & 50.17 & 30.00 \\
& MonoDETR~\cite{MonoDETR} & 51.33 & 37.46 & \textbf{61.40} & 31.61 & 19.36 \\
& MonoDGP~\cite{MonoDGP} & 54.08 & 39.92 & 57.07 & 28.52 & 34.16 \\
& MonoLAA(Ours) & \textbf{59.98} & \textbf{48.32} & 55.58 & \textbf{51.03} & \textbf{38.35} \\
\hline
\end{tabular}
}
\caption{3D object detection results on the LAA3D-real test set (detailed metrics and validation set results are reported in the Supp. Material). The best results are shown in bold, and the red ADS is the overall performance indicator. Columns 5 to 7 are 3D AP.}
\label{tab:LAA3D-real-test-set}
\end{table}

\begin{table}[htbp]
\centering
\small
\setlength{\tabcolsep}{4pt}
\resizebox{\linewidth}{!}{
\begin{tabular}{|l|l|>{\columncolor{blue!10}}c|c|c c c|}
\hline
Backbone & Method & \textcolor{red}{\textbf{ADS}} & mAP & MAV & eVTOL & Heli. \\
\hline
\multirow{3}{*}{ViT-L} 
& CenterNet~\cite{CenterNet} & 48.68 & 43.35 & 36.43 & 48.63 & \textbf{44.97} \\
& MonoDGP~\cite{MonoDGP} & 36.50 & 20.15 & 20.67 & 22.92 & 16.87 \\
& MonoLAA(Ours) & \textbf{50.74} & \textbf{45.92} & \textbf{38.57} & \textbf{55.06} & 44.13 \\
\hline
\multirow{3}{*}{ResNet152} 
& CenterNet~\cite{CenterNet} & 64.77 & 56.92 & 50.86 & 61.91 & 58.00 \\
& MonoDGP~\cite{MonoDGP} & 53.57 & 41.77 & 43.65 & 47.33 & 34.32 \\
& MonoLAA(Ours) & \textbf{71.18} & \textbf{68.10} & \textbf{59.13} & \textbf{75.05} & \textbf{70.11} \\
\hline
\multirow{3}{*}{ConvNeXt-L} 
& CenterNet~\cite{CenterNet} & 62.40 & 53.16 & 50.51 & 59.20 & 49.78 \\
& MonoDGP~\cite{MonoDGP} & 63.36 & 52.97 & \textbf{55.53} & 58.12 & 45.26 \\
& MonoLAA(Ours) & \textbf{64.64} & \textbf{58.56} & 55.06 & \textbf{65.15} & \textbf{55.48} \\
\hline
\end{tabular}
}
\caption{3D object detection results on the LAA3D-sim test set (detailed metrics and validation set results are reported in the Supp. Material). Columns 5 to 7 are 3D AP. }
\label{tab:LAA3D-sim-test-set}
\end{table}

\subsubsection{Main Results}

\textbf{Implementation details.}
We trained all detectors on 8$\times$4090 GPUs with Adam~\cite{Adam} optimizer. We adopted a learning rate of 0.001 and trained the network for 30 epochs on LAA3D-sim and 40 epochs on LAA3D-real. We employ the traditionally used data augmentation, including scaling~\cite{Pointrcnn}, flipping~\cite{Pointrcnn}, and photometric augmentation~\cite{dataaug}.

\textbf{Benchmark results.} 
We implemented four previous methods: CenterNet~\cite{CenterNet}, Center-DORN~\cite{DORN}, MonoDETR~\cite{MonoDETR}, and MonoDGP~\cite{MonoDGP}, where Center-DORN extends CenterNet~\cite{CenterNet} with bin-based depth regression~\cite{DORN}. All detectors are adapted from 4-DoF (x, y, z, and yaw) to full 6-DoF (x, y, z, roll, pitch, and yaw) estimation.
Tables~\ref{tab:LAA3D-real-test-set} and~\ref{tab:LAA3D-sim-test-set} report detection performance on the LAA3D-real and LAA3D-sim test sets using ResNet152~\cite{he2016deep}, ViT-L~\cite{dosovitskiy2020image}, and ConvNeXt-L~\cite{liu2022convnet} backbones with ImageNet pretraining. Overall, MonoLAA consistently outperforms all baselines across backbones and datasets.
Specifically, on LAA3D-real, MonoLAA improves ADS by 2–6 points over the strongest baseline across all backbones. On LAA3D-sim, MonoLAA achieves similar gains.
These results validate the effectiveness of our FLU and CSD modules.

\subsubsection{Ablation Study}
We conducted the ablation study on the LAA3D-real validation set using MonoLAA with ConvNeXt backbone.

\textbf{Components analysis.} 
Table~\ref{tab:monolaa_results_reorder} shows the impact of FLU and CSD. Adding FLU to the baseline improves ADS (+4.71) and mAP (+5.55), especially for eVTOL, demonstrating the benefit of focal-length alignment. Adding CSD further boosts ADS to 68.72, confirming that class-specific depth effectively handles heterogeneous object distances.

\begin{table}[htbp]
\centering
\small
\setlength{\tabcolsep}{4pt}
\resizebox{\linewidth}{!}{
\begin{tabular}{|l|c c|c |c|c c c|}
\hline
Method & FLU & CSD & ADS & mAP & MAV & eVTOL & Heli. \\
\hline
MonoLAA* &  &  & 57.91 & 45.37 & 56.03 & 43.31 & 36.78 \\
MonoLAA  & \checkmark &  & 62.62 & 50.92 & 59.62 & 55.25 & 37.89 \\
MonoLAA  & \checkmark & \checkmark & 68.72 & 59.50 & 64.69 & 60.03 & 53.79 \\
\hline
\end{tabular}
}
\caption{Component analysis results. * denotes inclusion of focal length as an extra input. }
\label{tab:monolaa_results_reorder}
\end{table}

\begin{table}[htbp]
\centering
\small
\setlength{\tabcolsep}{6pt}
\resizebox{\linewidth}{!}{
\begin{tabular}{|l|c c c|c | c|}
\hline
Rot. Param. & Scale & Flip & Photo Aug. & ADS & mAP \\
\hline
Quaternion &  &  &  & 65.88 & 57.32 \\
\hline
Sin-Cos &  &  &  & 67.08 & 58.83 \\
\hline
Sin-Cos & \checkmark &  &  & 67.90 & 58.87 \\
\hline
Sin-Cos &  & \checkmark &  & 68.68 & 59.34 \\
\hline
Sin-Cos &  &  & \checkmark & 67.46 & 57.96 \\
\hline
Sin-Cos & \checkmark & \checkmark & \checkmark & 68.97 & 59.50 \\
\hline
\end{tabular}
}
\caption{Rotation encoder and data augmentation analysis results.}
\label{tab:rotation_ablation}
\end{table}

\textbf{Data augmentation analysis.}
Table~\ref{tab:rotation_ablation} evaluates the effects of scaling, flipping, and photometric augmentation. Individually, crop-scale and flip improve ADS and mAP, with flip yielding the largest single gain (+1.6 ADS, +0.51 mAP). Photometric augmentation provides minor improvements. Combining all three augmentations further boosts performance to ADS 68.97 and mAP 59.50, demonstrating that these augmentations complement each other and enhance model robustness.

\textbf{Impact of rotation encoder.}
Table~\ref{tab:rotation_ablation} also compares rotation representations. Replacing quaternion with sin–cos encoding improves ADS from 65.88 to 67.08 and mAP from 57.32 to 58.83, indicating that the sin–cos formulation provides a more stable and accurate rotation representation for pose estimation.

\begin{table}[htbp]
\centering
\small
\setlength{\tabcolsep}{6pt}
\renewcommand{\arraystretch}{1.15}
\resizebox{\linewidth}{!}{
\begin{tabular}{|l|c c c|c |c|}
\hline
Input Size & MAV & eVTOL & Helicopter & ADS & mAP \\
\hline
1024$\times$512 & 71.74 & 62.31 & 44.44 & 68.97 & 59.50 \\
\hline
512$\times$512  & 66.66 & 54.12 & 52.03 & 66.42 & 57.61 \\
\hline
512$\times$256  & 64.45 & 52.69 & 47.33 & 63.57 & 54.82 \\
\hline
256$\times$256  & 60.30 & 49.27 & 51.92 & 62.23 & 53.83 \\
\hline
\end{tabular}
}
\caption{Ablation study on the effect of input image size.}
\label{tab:input_size_ablation}
\end{table}

\begin{table}[htbp]
\centering
\small
\setlength{\tabcolsep}{7pt}
\renewcommand{\arraystretch}{1.15}
\resizebox{\linewidth}{!}{
\begin{tabular}{|l|c |c c c c|}
\hline
\multirow{2}{*}{Input Size} & \multirow{2}{*}{ADS} & \multicolumn{4}{c|}{Run Time (ms)} \\
\cline{3-6}
 &  & Data Read & To GPU & Detect & All \\
\hline
1024$\times$512 & 68.97 & 29 & 52 & 26 & 107 \\
\hline
512$\times$512  & 66.42 & 29 & 40 & 18 & 87 \\
\hline
512$\times$256  & 63.57 & 29 & 29 & 15 & 73 \\
\hline
256$\times$256  & 62.23 & 29 & 20 & 13 & 62 \\
\hline
\end{tabular}
}
\caption{Effect of input image size on performance and runtime.}
\label{tab:input_size_runtime}
\end{table}

\begin{figure}
  \centering
  \includegraphics[width=0.48\textwidth]{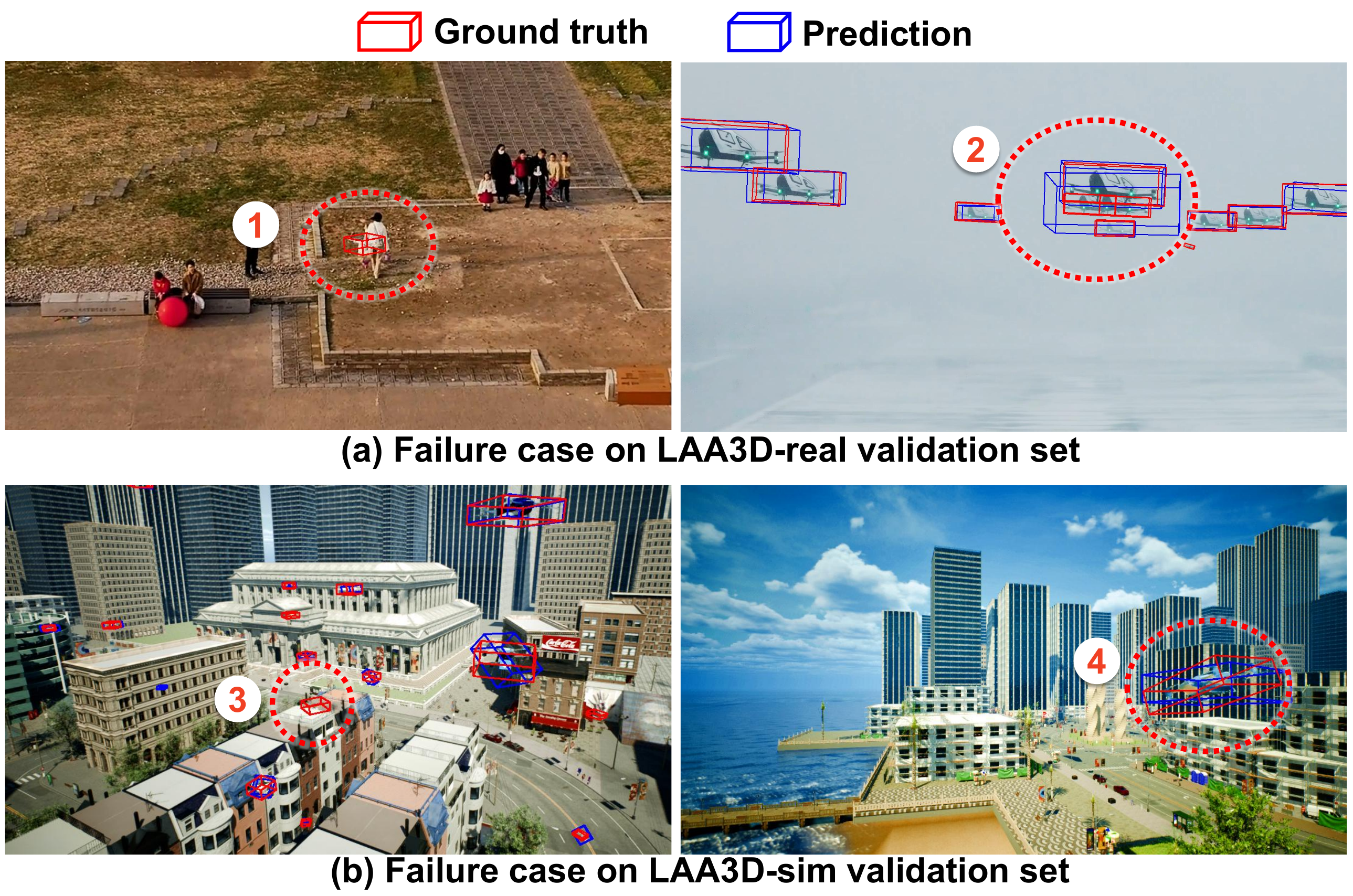}
  \caption{Failure cases on the LAA3D-real validation set.}
  \label{det_vi}
\end{figure}

\begin{table*}[htbp]
\centering
\small
\renewcommand{\arraystretch}{1.15}
\setlength{\tabcolsep}{4pt}
\resizebox{\textwidth}{!}{
\begin{tabular}{|l|l|
c c c c|
c c c c|
c c c c|}
\hline
\multirow{2}{*}{Tracker} & \multirow{2}{*}{Detector} &
\multicolumn{4}{c|}{MAV} &
\multicolumn{4}{c|}{eVTOL} &
\multicolumn{4}{c|}{Helicopter} \\
\cline{3-14}
& & HOTA↑ & MOTA↑ & MOTP↓ & MODA↑
  & HOTA↑ & MOTA↑ & MOTP↓ & MODA↑
  & HOTA↑ & MOTA↑ & MOTP↓ & MODA↑ \\
\hline
\multirow{5}{*}{AB3DMOT~\cite{AB3DMOT}} 
& CenterNet~\cite{CenterNet} &  47.67 & 37.62 & 1.47  & 39.44 & 10.71 & 14.85 & 2.57  & 15.71 & 1.01  & 7.27  & 4.36  & 8.36 \\
\cline{2-14}
& Center-DORN~\cite{DORN} & 54.03 & 49.93 & 1.34  & 51.75 & 11.04 & 33.87 & 1.34  & 34.38 & 1.16  & 13.09 & 4.21  & 14.18 \\
\cline{2-14}
& MonoDETR~\cite{MonoDETR} & 43.23 & 38.16 & 1.29  & 40.42 & 7.75  & -22.04 & 2.87  & -21.47 & 1.34  & -13.82 & 5.60   & -12.00 \\
\cline{2-14}
& MonoDGP~\cite{MonoDGP} &  53.80  & 40.82 & 1.29  & 42.17 & 10.40  & 4.11  & 2.72  & 4.23  & 3.10   & -7.64 & 4.77  & -4.73 \\
\cline{2-14}
& MonoLAA (Ours) & \textbf{56.22} & \textbf{57.03} & \textbf{1.20} & \textbf{59.10} & \textbf{11.39} & \textbf{35.24} & \textbf{1.31} & \textbf{36.09} & \textbf{3.46} & \textbf{35.64} & \textbf{3.26} & \textbf{36.73} \\
\hline
\end{tabular}}
\caption{
MOT results on the LAA3D-real test set (detailed metrics and validation set results are provided in the Supp. Material). 
}
\label{tab:tracking_results_resnet152}
\end{table*}

\textbf{Impact of input size.}
Tables~\ref{tab:input_size_ablation} and~\ref{tab:input_size_runtime} evaluate the effect of input image resolution. Increasing input resolution improves performance, with ADS rising from 62.23 (256×256) to 68.97 (1024×512) and mAP from 53.83 to 59.50, especially for MAVs and eVTOLs. Higher resolutions incur longer runtime (62ms to 107ms), highlighting the trade-off between accuracy and efficiency.

\subsubsection{Failure Case Analysis}
To better understand the limitations of MonoLAA, we qualitatively analyze representative failure cases in Fig.~\ref{det_vi}. Two main failure modes are observed: missed detections of small objects \circled{1} \circled{3} and false positives with false poses \circled{2} \circled{4}. These examples highlight the challenges of low-altitude 3D detection, including limited visibility, diverse object poses, and ambiguous backgrounds, underscoring the need for enhanced spatial reasoning, robust feature encoding, and improved contextual modeling.

\subsection{3D Multi-Object Tracking}

\textbf{Metrics.}
We use CLEAR MOT metrics~\cite{CLEAR} and HOTA metrics~\cite{Hota} with a distance-based association, where the positives are defined by the distance thresholds of [4, 6, 12]m for MAV, eVTOL, and Helicopter, respectively.

\textbf{Baselines.}
We evaluate five monocular 3D detectors, i.e., CenterNet~\cite{CenterNet}, Center-DORN~\cite{DORN}, MonoDETR~\cite{MonoDETR}, MonoDGP~\cite{MonoDGP}, and MonoLAA, combined with the AB3DMOT~\cite{AB3DMOT} tracker to establish multi-object tracking baselines on the LAA3D-real dataset.

\textbf{Evaluation results.}
Table~\ref{tab:tracking_results_resnet152} shows that the MonoLAA-based tracker achieves the highest HOTA across all categories, confirming that stronger detection boosts tracking. Negative metrics in some baselines result from severe missed detections or identity switches. eVTOL and Helicopter tracking is lower due to greater distances and harder localization, indicating directions for future improvement.

\begin{table}[htbp]
\centering
\small
\setlength{\tabcolsep}{4pt}
\renewcommand{\arraystretch}{1.15}
\resizebox{\linewidth}{!}{
\begin{tabular}{|l|l|c | c|c c c|}
\hline
Setting & Method & ADS & mAP & MAV& eVTOL & Heli.\\
\hline
Supervised & MonoLAA & 55.23 & 43.77 & 49.43 & 47.53 & 34.34 \\
\hline
\multirow{3}{*}{UDA}
& Direct Test & 8.64 & 4.83 & 12.37 & 2.05 & 0.06 \\
& BG Replace~\cite{MAV6D} & 9.89 & 4.00 & 10.24 & 1.72 & 0.02 \\
& Pseudo-label~\cite{ST3D} & 11.44 & 6.42 & 14.70 & 4.57 & 0.00 \\
\hline
DA & Fine-tuning & \textbf{63.24} & \textbf{53.88} & \textbf{68.43} & \textbf{57.51} & \textbf{35.70} \\
\hline
\end{tabular}}
\caption{Domain adaptation results on the LAA3D-real test set (detailed metrics and validation set results are provided in the Supp. Material).}
\label{tab:da_result_test_reordered}
\end{table}

\subsection{Simulation to Reality Domain Adaptation}
Sim-to-real domain adaptation is crucial for bridging the gap between simulation and reality in 3D LAA detection. In this work, we explore two adaptation strategies: Unsupervised Domain Adaptation (UDA), which transfers knowledge without using labeled real data, and Domain Adaptation (DA) via fine-tuning, which leverages labeled real data to further reduce the sim-to-real gap.

\subsubsection{Unsupervised Domain Adaptation}

\textbf{Baselines.}
We evaluate several unsupervised domain adaptation (UDA) strategies to bridge the gap from simulation to real-world data. Specifically, (1) direct testing on real data without any adaptation, (2) background replacement~\cite{MAV6D} where synthetic backgrounds are substituted to mimic real scenes, and (3) pseudo-labeling~\cite{ST3D}, which generates target-domain labels from model predictions for self-training.

\textbf{Evaluation results.}
Table~\ref{tab:da_result_test_reordered} shows that all UDA methods yield only marginal improvements over direct testing, with ADS ranging from 8.64 to 11.44 and mAP from 4.83 to 6.42. The results reveal that naive adaptation strategies are insufficient to handle the substantial sim-to-real domain gap. Performance degradation is particularly severe for distant objects, such as Helicopters, where differences in scale, appearance, and photometric distribution between simulation and real data pose significant challenges. These findings highlight the limitations of existing vanilla UDA methods and the need for more effective transfer techniques.

\subsubsection{Sim-to-real Fine-tuning}

\textbf{Fine-tuning performance.}
Fine-tuning MonoLAA on real data substantially improves performance, achieving ADS 63.24 and mAP 53.88, with consistent gains across all categories (MAV 68.43, eVTOL 57.51, Helicopter 35.70). Remarkably, this even surpasses the model trained solely on real data. These results demonstrate that targeted sim-to-real fine-tuning not only mitigates domain shifts and enhances 3D detection accuracy but also highlights the value of large-scale LAA3D-sim as a cost-effective pretraining source for practical deployment.

\begin{figure}
  \centering
  \includegraphics[width=0.48\textwidth]{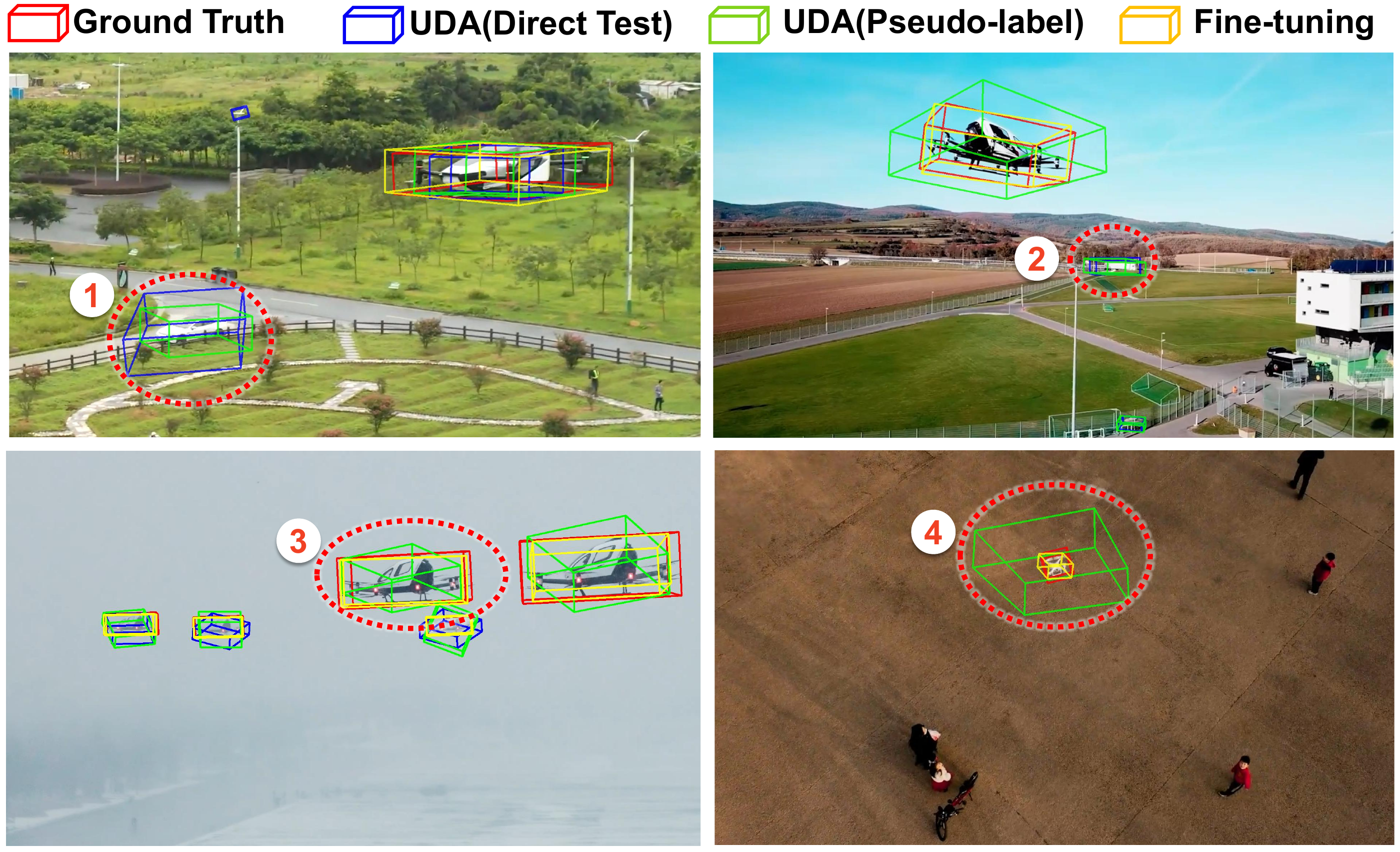}
  \caption{Qualitative results of sim-to-real domain adaptation. }
  \label{sim2real}
\end{figure}

\textbf{Qualitative Analysis.}
To visualize transfer performance, Fig.~\ref{sim2real} presents representative results. \circled{3} and \circled{4} show that pseudo-label-based UDA partially alleviates missed detections, but significant false positives remain in \circled{1} and \circled{2}, and pose estimation performance is poor. Fine-tuning on real data effectively addresses these issues, reducing false detections and improving pose estimation, which explains the substantial performance gains.

%% file: sec/5_conclu.tex
\section{Conclusion}
We present LAA3D, a large-scale dataset for low-altitude 3D aircraft perception, addressing the lack of high-quality annotated data for emerging aerial vehicles such as MAVs, eVTOLs, and helicopters. LAA3D combines real and synthetic images across diverse environments, weather, and illumination, with rich 3D annotations including 6-DoF poses, class labels, and instance identities. To benchmark this dataset, we propose MonoLAA, a monocular 3D detection baseline with Focal-Length Unification and Class-Specific Depth modules, achieving robust detection across varying focal lengths. Experiments show that models pretrained on LAA3D-sim transfer effectively to LAA3D-real via fine-tuning, demonstrating strong sim-to-real generalization and practical deployment potential.

%% file: sec/X_suppl.tex
\clearpage
\setcounter{page}{1}
\maketitlesupplementary

\section{Details of LAA3D Dataset }

\subsection{Details of LAA3D-real}

\subsubsection{Labeling Details. }

\begin{figure}
  \centering
  \includegraphics[width=0.45\textwidth]{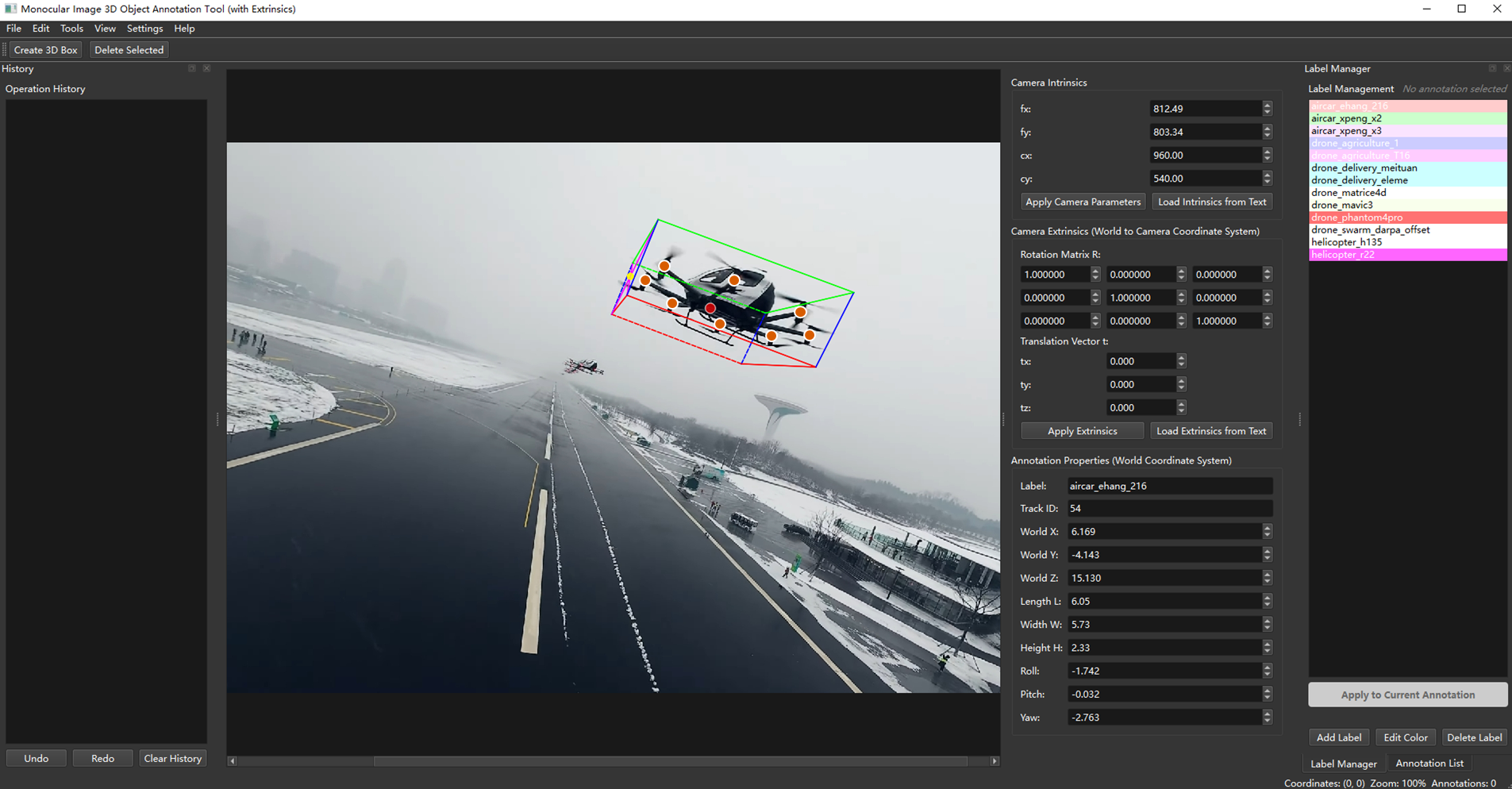}
  \caption{Labeling system.}
  \label{app_label_plantform}
\end{figure}

We employ a keypoint-based monocular 3D annotation system to address depth ambiguity inherent in single-view images (see Fig.~\ref{app_label_plantform}). The workflow ensures precise 6-DoF pose estimation and accurate 3D localization annotation.

\textbf{CAD model construction.} For each aircraft type (e.g., EHang 216), we build detailed CAD models with semantically meaningful keypoints (rotor hubs, fuselage endpoints) to serve as metric-accurate templates.

\textbf{Interactive 3D Box Manipulation.} Annotators adjust 3D bounding boxes with full 6-DoF control: (1)Translation: X, Y, Z position; (2)Rotation: Roll, pitch, yaw; (3)Dimensions: Length, width, height.

\textbf{Visual feedback.} Real-time rendering provides:
(1) Multi-color edges and depth-aware occlusion cues; (2) Keypoint markers for rapid spatial reference.

\textbf{Camera integration.} Annotation leverages camera calibration:
(1) Intrinsic matrices and extrinsic world-to-camera transformations; (2) Real-time 3D-to-2D projection to ensure visual alignment.

\textbf{Annotation workflow.} Systematic steps include:
(1) Load image and calibration; (2) Initialize CAD model; (3) Roughly position bounding box; (4) Fine-tune 6-DoF pose ensuring alignment of projected keypoints and box corners with image features, while maintaining physically plausible dimensions; (5) Assign track IDs for multi-object tracking.

\textbf{Quality assurance.} Each annotation is validated through:
(1) Automatic checks for invalid dimensions or poses; (2) Manual review by senior annotators; (3) Cross-frame verification for temporal coherence.

The dataset thus provides precise 6-DoF poses, accurate dimensions, class labels, and persistent track IDs, supporting 3D detection, tracking, and pose estimation.

\subsubsection{Data Split and Distribution Details}

\begin{figure*}
  \centering
  \includegraphics[width=0.90\textwidth]{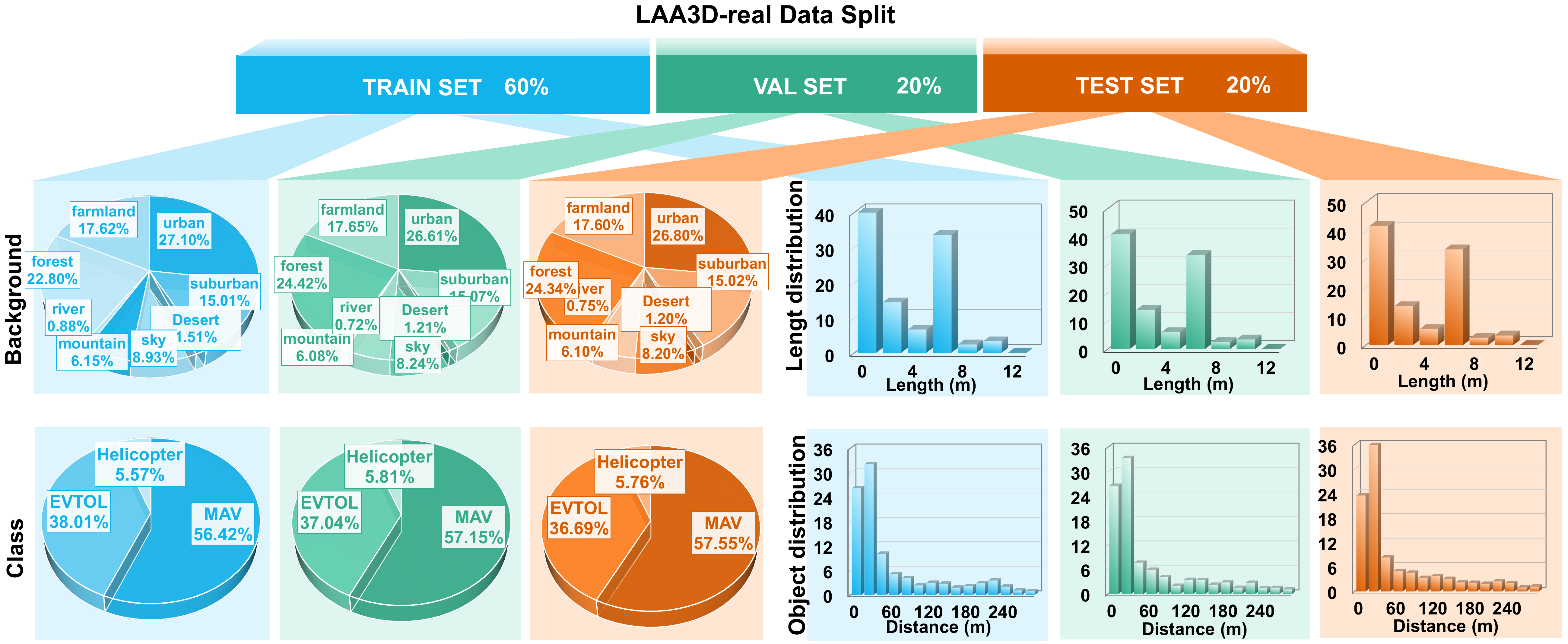}
  \caption{LAA3D-real data split statistics.}
  \label{real_split}
\end{figure*}

The LAA3D-real dataset is divided into non-overlapping training (60\%), validation (20\%), and test (20\%) subsets. This stratified split preserves the similar distributions of key attributes, i.e., object class, background, object length, and camera distance, ensuring fair and robust evaluation.
As shown in Fig.~\ref{real_split}, the dataset covers three categories: MAV, eVTOL, and Helicopter. Class ratios remain consistent across splits ($\sim$57\%, $\sim$37\%, and $\sim$6\%, respectively), preventing bias toward any type. Background distributions are also balanced, with urban ($\sim$27\%), forest ($\sim$23\%), farmland ($\sim$17\%), and suburban ($\sim$15\%) scenes dominating, while minor scenes (sky, mountain, river, desert) maintain proportional presence.
Object lengths (0–12m) and distances (0–300m) exhibit nearly identical histograms across splits, with most instances under 4m and within 100m of the camera.
Overall, the split design ensures statistical consistency across critical factors, providing a reliable benchmark for 3D detection evaluation.

\begin{figure*}
  \centering
  \includegraphics[width=0.95\textwidth]{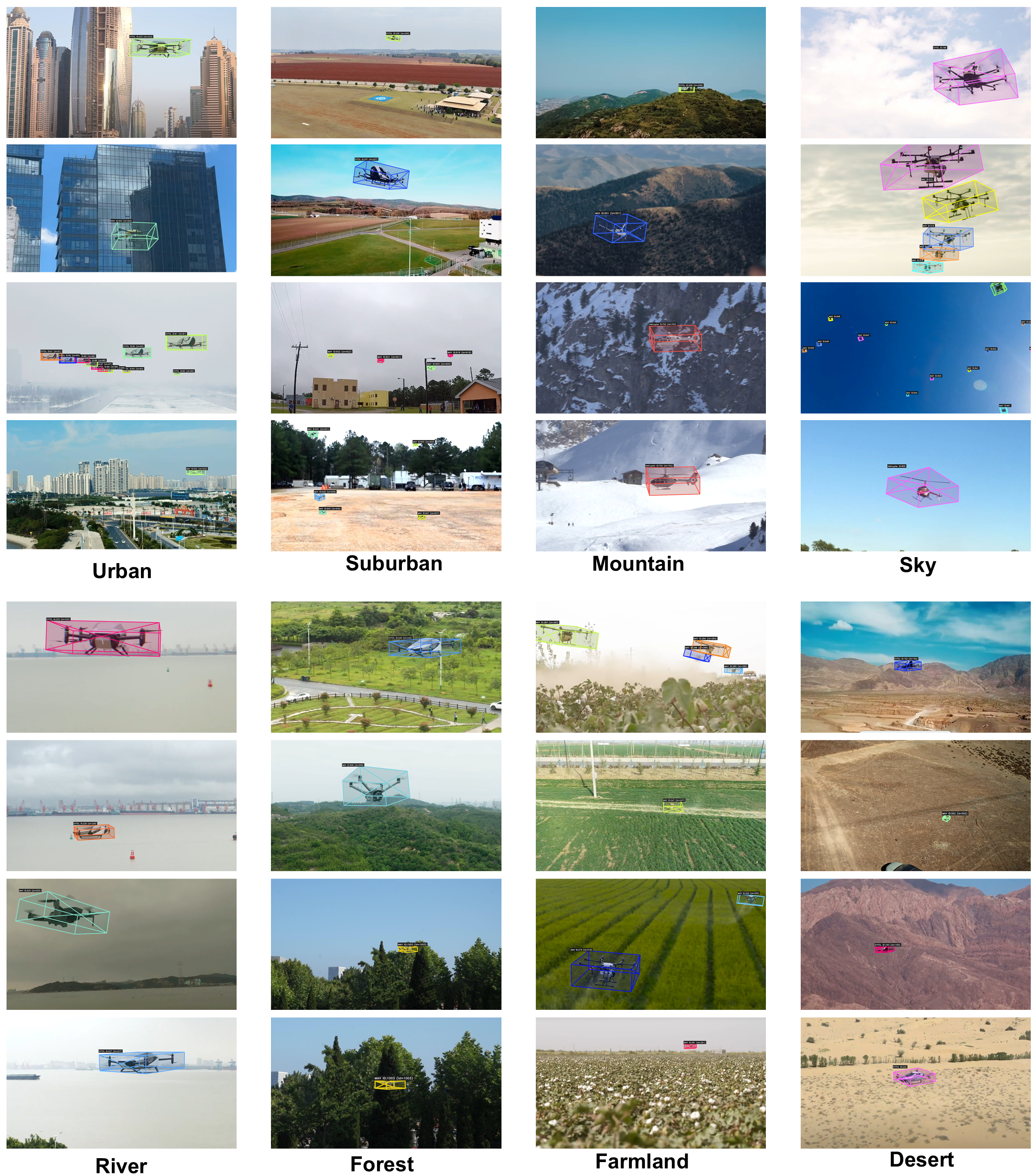}
  \caption{More examples from LAA3D-real dataset.}
  \label{app_real}
\end{figure*}

\subsubsection{More Examples}
To highlight the environmental diversity of LAA3D-real, Fig.~\ref{app_real} shows examples spanning eight scene types: Urban, Suburban, Mountain, Sky, River, Forest, Farmland, and Desert. Each includes multiple aerial agents (e.g., drones, eVTOLs) with precise 3D annotations, covering contexts from dense cities to remote natural terrains.
These scenes exhibit broad variations in terrain, clutter and agent density. Urban scenes show buildings. Farmlands contain structured textures, and deserts present open spaces. Sky, River, and Forest scenes further challenge perception with low-texture, and cluttered backgrounds.
Such diversity makes LAA3D-real a rigorous benchmark for testing the robustness and generalization of 3D perception models in real-world aerial environments.

\subsection{Details of LAA3D-sim }
\begin{figure*}
  \centering
  \includegraphics[width=0.95\textwidth]{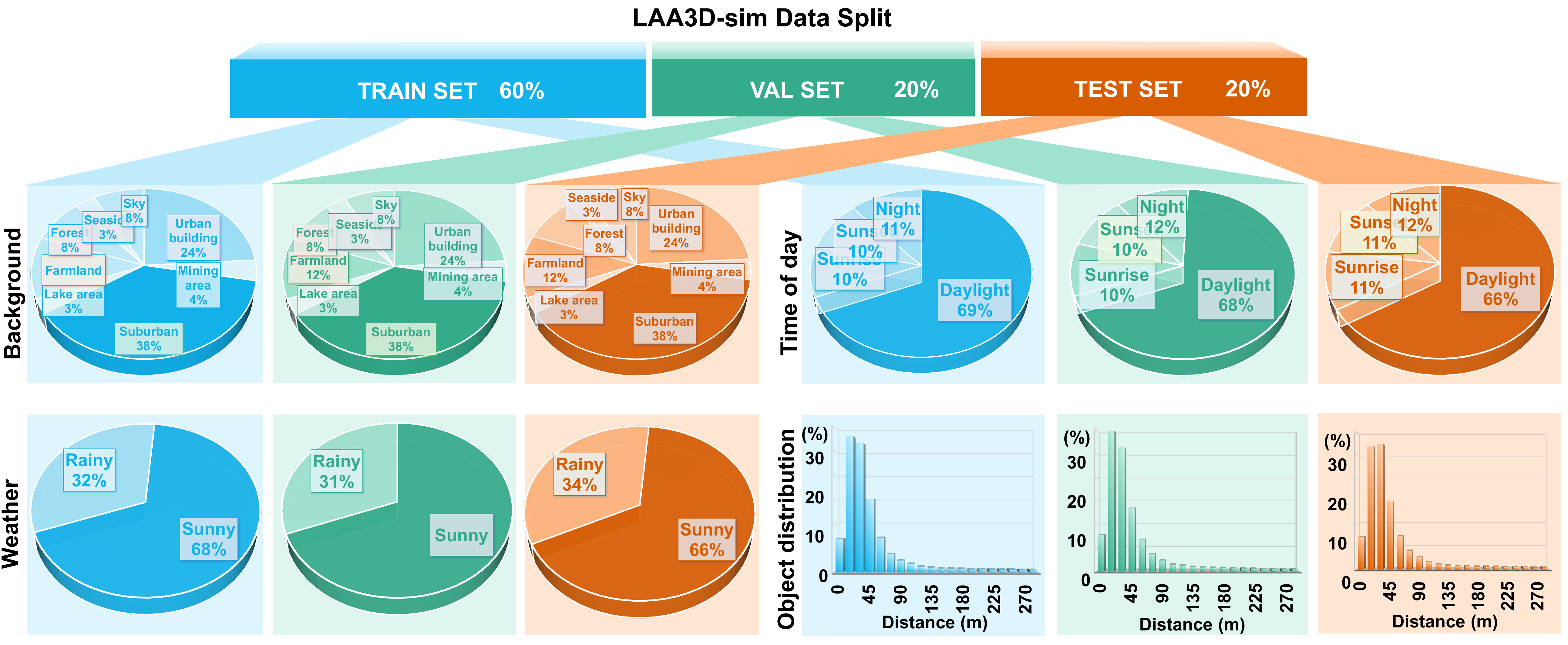}
  \caption{LAA3D-sim data split statistics.}
  \label{sim_split}
\end{figure*}
\subsubsection{Data Split and Distribution Details. }

To ensure fair and reproducible benchmarking, LAA3D-sim is divided into training (60\%), validation (20\%), and test (20\%) sets, preserving statistical consistency of key environmental and geometric attributes across all subsets (Fig.~\ref{sim_split}).
Background types, time-of-day, and weather distributions remain balanced, with variations within 1–2\%. Object–camera distance distributions are also consistent, with most objects within 0–100m, matching real-world aerial ranges.
Such consistency ensures that any synthetic-to-real performance gap arises from domain shift rather than dataset bias, establishing LAA3D-sim as a reliable benchmark for generalization and transfer in 3D aerial perception.

\begin{figure*}
  \centering
  \includegraphics[width=0.95\textwidth]{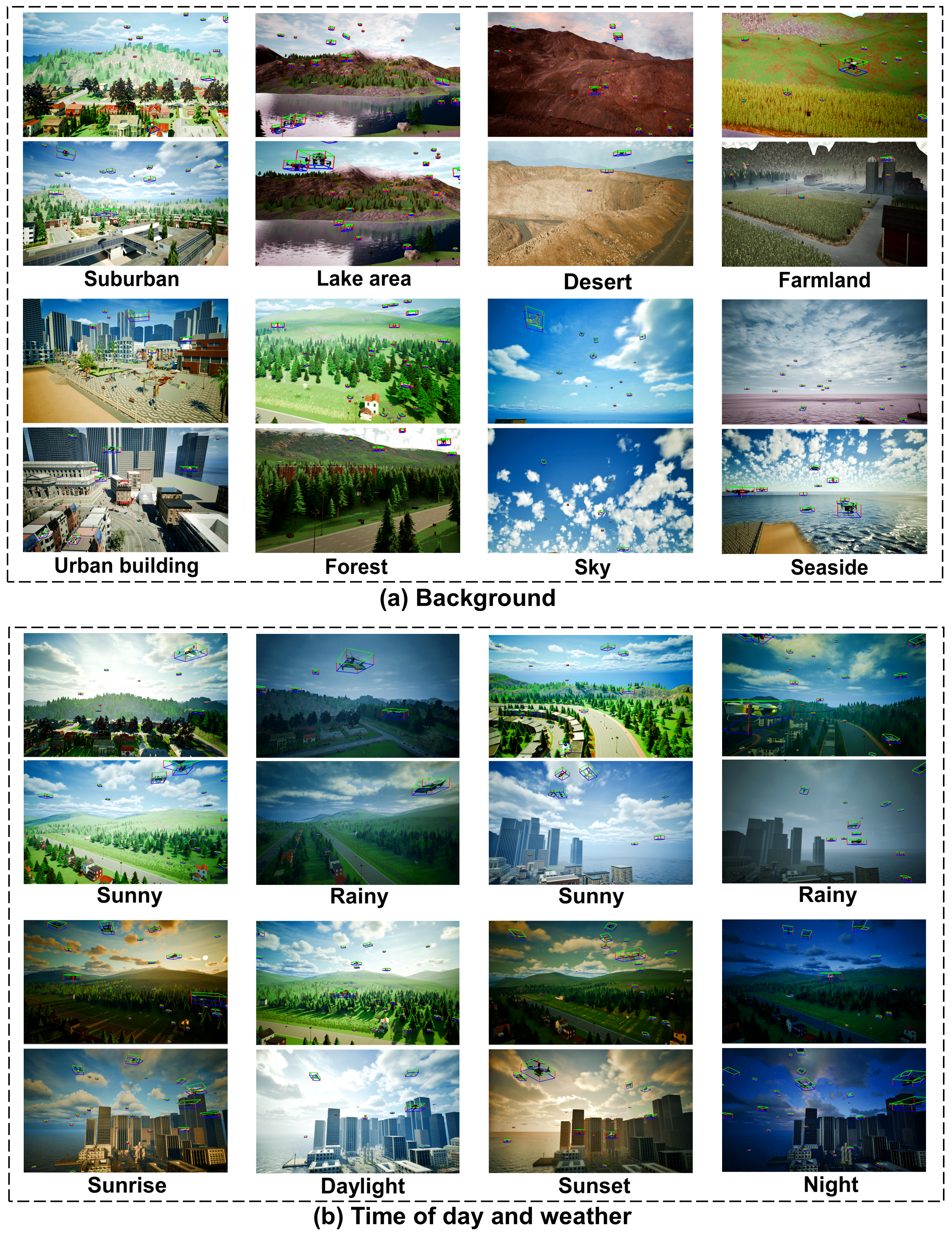}
  \caption{More examples from LAA3D-sim dataset.}
  \label{app_sim}
\end{figure*}

\subsubsection{More Examples }

To highlight the environmental and perceptual diversity of LAA3D-sim, Fig.~\ref{app_sim} presents representative examples across varied terrains and atmospheric conditions.
Fig.~\ref{app_sim}(a) covers eight background types, each containing multiple annotated aerial agents (e.g., drones, eVTOLs) rendered with realistic lighting and occlusion, capturing the complexity of multi-object perception.
Fig.~\ref{app_sim}(b) demonstrates diverse illumination and weather enabled by CARLA’s dynamic sky engine, including sunny, rainy, sunrise, sunset, and nighttime conditions with corresponding visual effects.
These variations ensure photometric and semantic realism, making LAA3D-sim a robust synthetic benchmark for training and evaluating 3D perception systems.

\section{Detailed Experiments and Analysis }

\subsection{3D Object Detection}
\subsubsection{Details of ADS metric.}
To address the limitations of conventional 3D detection metrics (e.g., 3D AP based on IoU) for small, distant aerial objects, which often yield zero IoU even under reasonable predictions, we propose the Aircraft Detection Score (ADS), a distance-based evaluation metric inspired by the NuScenes Detection Score (NDS)~\cite{nuscenes}, but specifically adapted to the unique characteristics of low-altitude aerial perception.

The ADS integrates three key components: mean Average Precision (mAP), true-positive error metrics (translation, rotation, size), and detection recall (mDR), weighted to balance precision, accuracy, and completeness:

{\footnotesize
\begin{equation}
    ADS = \frac{1}{8}\left(4mAP + 100\sum_{mTP\in\mathbb{T}\mathbb{P}}(1 - \mathcal{N}(mTP)) + mDR\right),
\end{equation}
}
where $mAP$ is the average AP across all classes. $mTP$ is the true positive mean error. $\mathbb{T}\mathbb{P} = \{m\Delta_T, m\Delta_R, m\Delta_S\}$ represents the mean translation (ATE), rotation (AOE), and size (ASE) errors over true positives.
$\mathcal{N}(\cdot)$ denotes min-max normalization using dataset-specific maximum error bounds.
$mDR$ is the mean detection recall rate, emphasizing the importance of not missing small or distant objects.
Below, we detail each component.

\textbf{3D AP Calculation.}
We compute 3D AP using distance-based matching rather than 3D IoU. For each class, we evaluate AP at multiple Euclidean distance thresholds. Since the evaluation distances for MAV, eVTOL, and Helicopter are approximately 2$\times$, 3$\times$, and 6$\times$ greater than those in nuScenes, we set the distance thresholds to [1, 2, 4, 8], [1.5, 3, 6, 12], and [3, 6, 12, 24] meters, respectively. The final 3D AP for a class is averaged over these thresholds:
\begin{equation}
    AP =  100\frac{1}{D_n} \sum_{i} AP_{\text{dis}=i},
\end{equation}
where $D_n = 4$ is the number of distance thresholds used per class. 

\textbf{Average translation error (ATE in meters).}
A prediction is considered a true positive if its projected 2D bounding box overlaps with the ground truth by an IoU of at least 0.1 on the image plane.
For each true positive prediction, the translation error is computed as the Euclidean distance between predicted center $\mathbf{p}_j$ and ground-truth center $\mathbf{p}'_j$:
\begin{equation}
    m\Delta_T = \frac{1}{N_p} \sum_j \|\mathbf{p}_j - \mathbf{p}'_j\|_2.
\end{equation}
Where $N_p$ is the true positive number. The $m\Delta_T$ is the average translation error among all classes.

\textbf{ Average rotation error (AOE in degrees).}
To handle orientation ambiguity (e.g., symmetric aircraft), we compute the minimal angular difference for each Euler angle $\theta \in \boldsymbol{\alpha}$ (predicted) and $\theta' \in \boldsymbol{\alpha}'$ (ground truth):
\begin{equation}
    \Delta \theta =  \min\left(|\theta - \theta'|,\, |\theta - (\theta' + \pi)|\right).
\end{equation}
The final rotation error is the mean absolute error over roll, pitch, and yaw:
\begin{equation}
   m\Delta_R = \frac{1}{N_p} \sum_j \frac{1}{3} \sum_{i} |\Delta \theta_{i,j}|.
\end{equation}
The $m\Delta_R$ is the average rotation error among all classes.

\textbf{ Average size error (ASE in meters).}
The size error measures the mean absolute deviation in  $s_i$(predicted length, width, and height) and $s'_i$ (ground truth):
\begin{equation}
    m\Delta_S = \frac{1}{N_p} \sum_j\sum_i\frac{1}{3} \|\mathbf{s}_{i,j} - \mathbf{s}'_{i,j}\|_1.
\end{equation}
The $m\Delta_S$ is the average size error among all classes.

\textbf{Normalization.}
Each $mTP$ component is normalized using class-specific maximum error values: for MAV, eVTOL, and Helicopter, we set max translation errors to [4, 6, 12] m, max rotation errors to [45°, 45°, 45°], and max size errors to [0.5, 0.5, 0.5] m.

Finally, the inclusion of $mDR$(\%), defined as the ratio of true positives to total ground truths, ensures that models are penalized for missed detections, which are particularly prevalent for small or distant aerial objects.

This comprehensive formulation makes ADS a more holistic and realistic metric for evaluating 3D perception systems in low-altitude aerial domains, balancing localization accuracy, geometric fidelity, and detection completeness.

\subsubsection{3D Object Detection Details}
Due to space constraints in the main paper, we provide comprehensive 3D object detection results in this supplementary material. Specifically, we include detailed evaluations on both the validation and test sets of LAA3D-real and LAA3D-sim. The results cover (1) overall performance comparison tables, (2) class-wise 3D Average Precision 3DAP metrics, and (3) precision-recall (PR) curves for each category. 

For ease of reference, we summarize the corresponding tables and figures below:

\begin{table}[h]
\centering
\small
\setlength{\tabcolsep}{6pt}
\renewcommand{\arraystretch}{1.1}
\resizebox{\linewidth}{!}{
\begin{tabular}{l|ccc}
\hline
Split / Data & Metric & LAA3D-real & LAA3D-sim \\
\hline
\multirow{3}{*}{\textit{Test set}}
& Overall & Table~\ref{full_real_test_results} & Table~\ref{full_sim_test_results} \\
& 3DAP & Table~\ref{full_real_test_ap} & Table~\ref{full_sim_test_ap} \\
& 3DAP Curve & Fig.~\ref{app_real_ap} & Fig.~\ref{app_sim_ap} \\
\hline
\multirow{3}{*}{\textit{Val. set}}
& Overall & Table~\ref{full_real_val_results} & Table~\ref{full_sim_val_results} \\
& 3DAP & Table~\ref{full_real_val_ap} & Table~\ref{full_sim_val_ap} \\
& 3DAP Curve & Fig.~\ref{app_real_ap_val} & Fig.~\ref{app_sim_ap_val} \\
\hline
\end{tabular}}
\caption{Summary of detailed 3D detection results provided in the supplementary material.}
\end{table}

These results provide a complete view of model performance across domains and data splits, complementing the summary results reported in the main paper.

\subsection{Domain Adaptation Details}

Due to space constraints in the main paper, we provide additional detailed results on domain adaptation experiments conducted on both the LAA3D-real validation and test sets. These extended tables (Table~\ref{full_val_da} and Table~\ref{full_test_da}) offer a fine-grained breakdown of performance across individual object classes (MAV, eVTOL, Helicopter) as well as overall metrics, enabling deeper analysis of model behavior under domain shift.

The detailed experimental data further demonstrate that:
(1) Supervised training (MonoLAA) achieves strong performance when trained and evaluated on the same domain, but suffers significant degradation under unsupervised domain adaptation (UDA) settings, confirming the substantial simulation-to-real gap in low-altitude aerial perception.
(2) Simple UDA baselines such as Direct Test, BG Replace~\cite{MAV6D}, and Pseudo-label~\cite{ST3D} show limited effectiveness. While pseudo-labeling improves detection recall (mDR) and 3D AP for MAVs, it fails to generalize to larger aircraft like helicopters, where detection rates remain near zero. This highlights the challenge of adapting to diverse object scales without explicit supervision.
(3)In contrast, fine-tuning on real-world data yields substantial gains across all metrics, indicating that domain-adapted models are not only more accurate but also more complete and robust.

\subsection{6-DoF Pose Estimation Details}
While 3D object detection jointly evaluates class, size, and 6-DoF pose, 6-DoF pose estimation specifically targets the accuracy of predicted orientation and translation. This task is commonly evaluated using the Average Distance (ADD) and Symmetric Average Distance (ADD-S) metrics, which measure the mean distance between transformed model points under predicted and ground-truth poses. To facilitate research on aerial 6-DoF pose estimation, we report baseline results on the LAA3D-real validation set (Table~\ref{pose_val}) using three representative methods: Yolo-6D~\cite{YOLO6D}, YOLOX-6D-Pose~\cite{Yolo-6d-pose}, and YOLOV5-6D~\cite{YOLOV5-6D}. We report the percentage of instances for which the ADD or ADD-S error falls below 50\% of the object’s diameter, i.e., 50\% ADD/ADD-S, reflecting the proportion of predictions with geometrically acceptable pose alignment.

The methods can be broadly categorized into two paradigms:
(1) PnP-based approaches (Yolo-6D, YOLOV5-6D), which first detect 2D keypoints or bounding boxes in the image and then recover 6-DoF pose via Perspective-n-Point solvers.
(2) Direct regression approaches (YOLOX-6D-Pose), which predict 6-DoF parameters end-to-end from image features without intermediate geometric steps. 

These results establish the benchmark for 6-DoF pose estimation on real-world low-altitude aerial scenes. They also underscore the complementary value of ADD/ADD-S metrics in assessing fine-grained geometric fidelity.

\subsection{3D MOT Details}

Due to space limitations, we present detailed 3D MOT results on LAA3D-real validation and test sets, reporting three metric families: (1) CLEAR MOT metrics (MOTA, MOTP, MODA, IDSWs)~\cite{CLEAR}; (2) identity metrics (IDF1, IDTP, IDFP, IDFN); and (3) HOTA metrics (HOTA, DetA, AssA)~\cite{Hota}, with HOTA providing a comprehensive evaluation of detection and association quality. The metrics is calculated by a distance-based association, where the positives are defined by the distance thresholds of [4, 6, 12]m for MAV, eVTOL, and Helicopter, respectively.

We construct five trackers based on the detections of baseline detectors and AB3DMOT~\cite{AB3DMOT} tracker. The AB3DMOT uses distance-based association.
Results in Tables~\ref{mot_val} and \ref{mot_test} show that the MonoLAA-based tracker consistently outperforms alternatives across all object classes. On MAVs, it achieves MOTA of 76.25 (val) and 58.85 (test), with superior IDF1 (88.48/79.64). 

Negative MOTA/MODA values observed in some methods (e.g., MonoDETR on eVTOLs) stem from excessive false positives overwhelming true positives, underscoring the critical dependence of 3D MOT on detection quality. 

These results confirm MonoLAA's superiority not only in 3D detection but also as a foundation for robust multi-object tracking in diverse low-altitude aerial scenarios.

\subsection{Trajectory Prediction Details}

For completeness and to facilitate future research, we present two widely adopted baseline methods for trajectory prediction: Kalman Filter (KF)~\cite{KF} and Long Short-Term Memory (LSTM)~\cite{lstm}. We evaluate both models using the standard metrics Average Displacement Error (ADE) and Final Displacement Error (FDE), based on 3-frame historical trajectories to predict the next 10 frames. All results are reported on the LAA3D-real validation set.

The performance of the baselines is summarized in Table~\ref{tab:trajectory_baselines}. As expected, LSTM outperforms the Kalman Filter across all difficulty levels, demonstrating the benefit of learning nonlinear temporal dynamics from data. These baselines serve as a practical reference for evaluating more sophisticated trajectory prediction models in 3D low-altitude scenarios.

\begin{table}[h]
\centering
\footnotesize
\begin{tabular}{lcc}
\hline
\multirow{2}*{Method} & Easy & Hard \\
\cline{2-3}
 & ADE / FDE & ADE / FDE \\
\hline
Kalman Filter~\cite{KF} & 0.3737 / 0.7786 & 0.5507 / 1.1035 \\
LSTM~\cite{lstm} & 0.3038 / 0.6865 & 0.4914 / 1.0462 \\
\hline
\end{tabular}
\caption{Trajectory prediction performance (ADE / FDE in meters) on LAA3D-real validation set.}
\label{tab:trajectory_baselines}
\end{table}

\begin{table*}[htbp]
\centering
\small
\setlength{\tabcolsep}{2pt}
\resizebox{\textwidth}{!}{
\begin{tabular}{|l|l|ccccc|ccccc|ccccc|cccc|cc|}
\hline
\multirow{2}*{Backbone} & \multirow{2}*{Method} &
\multicolumn{5}{c|}{MAV} &
\multicolumn{5}{c|}{eVTOL} &
\multicolumn{5}{c|}{Helicopter} &
\multirow{2}*{mAOE↓} & \multirow{2}*{mATE↓} & \multirow{2}*{mASE↓} & \multirow{2}*{mDR↑} & \multirow{2}*{mAP↑} & \multirow{2}*{ADS↑} \\
\cline{3-17}
& & AOE↓ & ATE↓ & ASE↓ & DR↑ & 3DAP↑ &
AOE↓ & ATE↓ & ASE↓ & DR↑ & 3DAP↑ &
AOE↓ & ATE↓ & ASE↓ & DR↑ & 3DAP↑ &
 & & & & & \\
\hline
\multirow{5}{*}{ViT-L}
& CenterNet~\cite{CenterNet} & 20.18 & 3.85 & 4.04 & 78.75 & 35.20 & 5.28 & 6.98 & 3.88 & 83.59 & 36.70 & 8.30 & 7.93 & 15.40 & 52.17 & 24.65 & 11.25 & 6.25 & 7.77 & 71.50 & 32.18 & 46.53 \\
& Center-DORN~\cite{DORN} & 27.19 & 3.35 & 9.28 & 74.64 & 38.13 & 7.84 & 5.77 & 8.14 & 85.11 & 46.17 & 15.20 & 11.12 & 51.90 & 49.28 & 21.50 & 16.74 & 6.55 & 23.11 & 69.68 & 35.33 & 42.66 \\
& MonoDETR~\cite{MonoDETR} & 28.15 & 2.46 & 2.93 & 97.12 & 49.43 & 12.27 & 9.16 & 6.74 & 92.29 & 21.02 & 19.87 & 12.10 & 19.08 & 98.55 & 33.71 & 20.10 & 7.91 & 9.58 & 95.99 & 34.72 & 47.98 \\
& MonoDGP~\cite{MonoDGP} & 21.94 & 2.55 & 1.93 & 97.90 & 52.03 & 8.00 & 7.17 & 1.05 & 93.26 & 27.59 & 19.92 & 25.94 & 6.14 & 99.27 & 12.40 & 16.62 & 11.89 & 3.04 & 96.81 & 30.67 & 48.58 \\
& MonoLAA(Ours) & 24.72 & 2.22 & 4.02 & 82.03 & 45.16 & 8.02 & 5.77 & 4.28 & 84.92 & 46.25 & 12.05 & 6.57 & 27.26 & 54.35 & 34.44 & 14.93 & 4.85 & 11.85 & 73.77 & 41.95 & 51.46 \\
\hline
\multirow{5}{*}{ResNet152}
& CenterNet~\cite{CenterNet} & 16.31 & 3.44 & 5.45 & 91.17 & 43.09 & 5.85 & 5.76 & 3.64 & 90.46 & 37.63 & 10.17 & 10.67 & 14.84 & 60.87 & 29.24 & 10.78 & 6.62 & 7.98 & 80.83 & 36.65 & 49.65 \\
&Center-DORN~\cite{DORN} & 23.64 & 2.73 & 12.98 & 89.11 & 47.19 & 9.28 & 6.03 & 11.39 & 88.55 & 44.75 & 10.47 & 6.24 & 25.98 & 49.28 & 30.56 & 14.46 & 5.00 & 16.78 & 75.64 & 40.83 & 49.98 \\
& MonoDETR~\cite{MonoDETR} & 22.22 & 2.48 & 4.17 & 95.75 & 49.02 & 8.46 & 9.59 & 12.58 & 93.28 & 22.57 & 23.71 & 17.47 & 34.82 & 99.25 & 16.02 & 18.13 & 9.85 & 17.19 & 96.09 & 29.20 & 43.86 \\
& MonoDGP~\cite{MonoDGP} & 23.05 & 2.78 & 2.35 & 94.62 & 47.08 & 6.08 & 6.78 & 0.89 & 97.49 & 31.94 & 18.95 & 18.73 & 7.22 & 97.20 & 26.53 & 16.03 & 9.43 & 3.49 & 96.43 & 35.18 & 50.71 \\
& MonoLAA(Ours) & 22.02 & 2.32 & 1.83 & 92.08 & 49.43 & 7.80 & 5.16 & 4.98 & 91.41 & 47.53 & 10.26 & 6.37 & 17.36 & 56.52 & 34.34 & 13.36 & 4.61 & 8.05 & 80.00 & 43.77 & 55.23 \\
\hline
\multirow{5}{*}{ConvNeXt-L}
&  CenterNet~\cite{CenterNet} & 17.18 & 2.39 & 3.63 & 92.38 & 51.14 & 5.31 & 5.66 & 3.25 & 92.94 & 39.28 & 6.59 & 7.64 & 11.11 & 57.97 & 30.18 & 9.69 & 5.23 & 5.99 & 81.10 & 40.20 & 54.48 \\
&Center-DORN~\cite{DORN} & 23.19 & 2.25 & 4.99 & 92.16 & 54.54 & 6.79 & 5.08 & 5.93 & 92.75 & 50.17 & 8.49 & 9.24 & 22.28 & 55.80 & 30.00 & 12.82 & 5.52 & 11.06 & 80.23 & 44.90 & 54.57 \\
& MonoDETR~\cite{MonoDETR} & 20.74 & 1.98 & 1.89 & 98.44 & 61.40 & 6.09 & 5.51 & 6.05 & 95.18 & 31.61 & 22.39 & 21.20 & 22.25 & 95.34 & 19.36 & 16.40 & 9.56 & 10.06 & 96.32 & 37.46 & 51.33 \\
& MonoDGP~\cite{MonoDGP} & 18.48 & 2.00 & 1.22 & 95.32 & 57.07 & 6.16 & 7.07 & 1.06 & 94.60 & 28.52 & 23.06 & 11.33 & 7.99 & 94.34 & 34.16 & 15.90 & 6.80 & 3.42 & 94.75 & 39.92 & 54.08 \\
& MonoLAA(Ours) & 19.38 & 1.72 & 1.66 & 92.38 & 55.58 & 5.91 & 4.14 & 3.19 & 90.08 & 51.03 & 8.32 & 5.66 & 14.76 & 57.97 & 38.35 & 11.21 & 3.84 & 6.54 & 80.14 & 48.32 & 59.98 \\
\hline
\end{tabular}
}
\caption{Detailed 3D object detection results on the LAA3D-real test set.}
\label{full_real_test_results}
\end{table*}

\begin{table*}[htbp]
\centering
\small
\setlength{\tabcolsep}{2pt}
\resizebox{\textwidth}{!}{
\begin{tabular}{|l|l|ccccc|ccccc|ccccc|cccc|cc|}
\hline
\multirow{2}*{Backbone} & \multirow{2}*{Method} &
\multicolumn{5}{c|}{MAV} &
\multicolumn{5}{c|}{eVTOL} &
\multicolumn{5}{c|}{Helicopter} &
\multirow{2}*{mAOE↓} & \multirow{2}*{mATE↓} & \multirow{2}*{mASE↓} & \multirow{2}*{mDR↑} & \multirow{2}*{mAP↑} & \multirow{2}*{ADS} \\
\cline{3-17}
& & AOE↓ & ATE↓ & ASE↓ & DR↑ & 3DAP↑ &
AOE↓ & ATE↓ & ASE↓ & DR↑ & 3DAP↑ &
AOE↓ & ATE↓ & ASE↓ & DR↑ & 3DAP↑ &
 & & & & & \\
\hline
\multirow{3}{*}{ViT-L}
& CenterNet~\cite{CenterNet} & 25.91 & 3.50 & 5.70 & 69.09 & 36.43 & 10.61 & 4.99 & 9.13 & 88.34 & 48.63 & 7.56 & 13.02 & 52.28 & 89.02 & 44.97 & 14.69 & 7.17 & 22.37 & 82.15 & 43.35 & 48.68 \\
& MonoDGP~\cite{MonoDGP} &30.53 & 4.54  & 5.51  & 78.91 & 20.67 & 14.89 & 8.34  & 9.84  & 93.91 & 22.92 & 11.17 & 19.29 & 33.8  & 85.34 & 16.87 & 18.86 & 10.72 & 16.38 & 86.05 & 20.15 & 36.50 \\
& MonoLAA(Ours) & 26.17 & 3.30 & 6.14 & 68.37 & 38.57 & 11.43 & 4.60 & 9.76 & 88.52 & 55.06 & 8.50 & 10.11 & 51.27 & 87.50 & 44.13 & 15.37 & 6.00 & 22.39 & 81.46 & 45.92 & 50.74 \\
\hline
\multirow{3}{*}{ResNet152}
& CenterNet~\cite{CenterNet} & 17.90 & 2.20 & 2.30 & 81.81 & 50.86 & 5.66 & 3.06 & 3.86 & 93.76 & 61.91 & 4.53 & 9.08 & 20.60 & 93.29 & 58.00 & 9.36 & 4.78 & 8.92 & 89.62 & 56.92 & 64.77 \\
& MonoDGP~\cite{MonoDGP} & 19.16 & 3.22 & 3.02 & 84.17 & 43.65 & 7.87 & 4.56 & 2.96 & 95.97 & 47.33 & 7.62 & 14.05 & 21.62 & 93.07 & 34.32 & 11.55 & 7.27 & 9.20 & 91.07 & 41.77 & 53.57 \\
& MonoLAA(Ours) & 18.02 & 1.79 & 2.62 & 79.51 & 59.13 & 6.19 & 2.50 & 4.37 & 94.21 & 75.05 & 5.39 & 7.75 & 22.51 & 93.29 & 70.11 & 9.87 & 4.01 & 9.83 & 89.01 & 68.10 & 71.18 \\
\hline
\multirow{3}{*}{ConvNeXt-L}
& CenterNet~\cite{CenterNet} & 17.58 & 2.40 & 2.48 & 86.89 & 50.51 & 6.77 & 3.58 & 5.53 & 94.93 & 59.20 & 6.03 & 9.33 & 17.23 & 93.27 & 49.78 & 10.13 & 5.10 & 8.41 & 91.70 & 53.16 & 62.40 \\
& MonoDGP~\cite{MonoDGP} & 15.74 & 2.20 & 1.26 & 89.93 & 55.53 & 5.24 & 3.29 & 2.19 & 96.71 & 58.12 & 5.39 & 10.53 & 14.79 & 91.09 & 45.26 & 8.79 & 5.34 & 6.08 & 92.58 & 52.97 & 63.36 \\
& MonoLAA(Ours) & 18.77 & 2.11 & 2.37 & 86.82 & 55.06 & 7.70 & 3.05 & 6.68 & 93.89 & 65.15 & 7.20 & 9.62 & 24.79 & 94.21 & 55.48 & 11.22 & 4.93 & 11.28 & 91.64 & 58.56 & 64.64 \\
\hline
\end{tabular}
}
\caption{Detailed 3D object detection results on the LAA3D-sim test set.}
\label{full_sim_test_results}
\end{table*}

\begin{table*}[htbp]
\centering
\small
\setlength{\tabcolsep}{6pt}

\resizebox{\textwidth}{!}{
\begin{tabular}{|l|l|ccccc|ccccc|ccccc|}
\hline
\multirow{2}*{Backbone} & \multirow{2}*{Method} &
\multicolumn{5}{c|}{MAV 3D AP↑} &
\multicolumn{5}{c|}{eVTOL 3D AP↑} &
\multicolumn{5}{c|}{Helicopter 3D AP↑} \\
\cline{3-17}
& & AP$_1$ & AP$_2$ & AP$_4$ & AP$_5$ & mAP &
AP$_{1.5}$ & AP$_3$ & AP$_6$ & AP$_{12}$ & mAP &
AP$_3$ & AP$_6$ & AP$_{12}$ & AP$_{24}$ & mAP \\
\hline
\multirow{5}{*}{ViT-L}
 & CenterNet~\cite{CenterNet} & 10.10 & 23.19 & 41.57 & 65.93 & 35.20 & 11.76 & 23.22 & 44.86 & 66.98 & 36.70 & 5.94 & 13.77 & 31.57 & 47.30 & 24.65 \\
& Center-DORN~\cite{DORN} & 19.24 & 26.79 & 47.50 & 58.98 & 38.13 & 25.00 & 37.19 & 57.28 & 65.99 & 46.17 & 5.10 & 11.70 & 31.43 & 37.78 & 21.50 \\
& MonoDETR~\cite{MonoDETR} & 12.14 & 32.37 & 67.95 & 85.27 & 49.43 & 1.83 & 6.55 & 22.87 & 52.82 & 21.02 & 9.54 & 19.50 & 42.42 & 63.38 & 33.71 \\
& MonoDGP~\cite{MonoDGP} & 15.86 & 35.06 & 68.96 & 88.22 & 52.03 & 3.31 & 10.61 & 30.00 & 66.46 & 27.59 & 1.76 & 6.17 & 13.11 & 28.58 & 12.40 \\
& MonoLAA (Ours) & 20.23 & 33.38 & 57.41 & 69.61 & 45.16 & 24.33 & 39.49 & 55.45 & 65.73 & 46.25 & 16.71 & 29.37 & 39.37 & 52.30 & 34.44 \\
\hline
\multirow{5}{*}{ResNet152}
 & CenterNet~\cite{CenterNet} & 11.44 & 27.74 & 54.28 & 78.91 & 43.09 & 6.72 & 20.21 & 48.53 & 75.05 & 37.63 & 8.27 & 18.32 & 35.58 & 54.80 & 29.24 \\
& Center-DORN~\cite{DORN} & 21.49 & 32.40 & 58.39 & 76.46 & 47.19 & 25.75 & 36.69 & 52.06 & 64.48 & 44.75 & 11.79 & 22.32 & 41.19 & 46.94 & 30.56 \\
& MonoDETR~\cite{MonoDETR} & 14.25 & 36.66 & 64.17 & 80.99 & 49.02 & 5.87 & 11.85 & 23.70 & 48.86 & 22.57 & 4.06 & 9.51 & 17.39 & 33.11 & 16.02 \\
& MonoDGP~\cite{MonoDGP} & 15.19 & 32.23 & 56.64 & 84.26 & 47.08 & 5.96 & 13.26 & 37.73 & 70.82 & 31.94 & 5.35 & 12.84 & 33.57 & 54.36 & 26.53 \\
& MonoLAA (Ours) & 18.54 & 35.04 & 63.43 & 80.72 & 49.43 & 23.70 & 38.39 & 56.14 & 71.89 & 47.53 & 13.33 & 28.65 & 42.90 & 52.48 & 34.34 \\
\hline
\multirow{5}{*}{ConvNeXt-L}
& CenterNet~\cite{CenterNet} & 15.44 & 33.26 & 69.36 & 86.51 & 51.14 & 8.63 & 24.77 & 50.34 & 73.39 & 39.28 & 4.65 & 18.06 & 43.84 & 54.17 & 30.18 \\
& Center-DORN~\cite{DORN} & 24.07 & 41.25 & 70.43 & 82.43 & 54.54 & 26.51 & 38.48 & 60.69 & 74.99 & 50.17 & 11.87 & 19.65 & 41.02 & 47.45 & 30.00 \\
& MonoDETR~\cite{MonoDETR} & 24.12 & 53.04 & 78.02 & 90.41 & 61.40 & 3.95 & 13.76 & 38.09 & 70.63 & 31.61 & 4.92 & 11.89 & 22.44 & 38.19 & 19.36 \\
& MonoDGP~\cite{MonoDGP} & 17.33 & 46.34 & 72.93 & 91.70 & 57.07 & 2.49 & 10.00 & 31.35 & 70.25 & 28.52 & 6.37 & 15.72 & 36.53 & 78.00 & 34.16 \\
& MonoLAA (Ours) & 23.66 & 43.39 & 70.70 & 84.59 & 55.58 & 22.61 & 41.54 & 61.49 & 78.48 & 51.03 & 22.30 & 34.15 & 40.80 & 56.16 & 38.35 \\
\hline
\end{tabular}
}
\caption{ Detailed 3D AP (\%) results on LAA3D-real test set.}
\label{full_real_test_ap}
\end{table*}

\begin{table*}[htbp]
\centering
\small
\setlength{\tabcolsep}{6pt}

\resizebox{\textwidth}{!}{
\begin{tabular}{|l|l|ccccc|ccccc|ccccc|}
\hline
\multirow{2}*{Backbone} & \multirow{2}*{Method} &
\multicolumn{5}{c|}{MAV 3D AP↑} &
\multicolumn{5}{c|}{eVTOL 3D AP↑} &
\multicolumn{5}{c|}{Helicopter 3D AP↑} \\
\cline{3-17}
& & AP$_1$ & AP$_2$ & AP$_4$ & AP$_5$ & mAP &
AP$_{1.5}$ & AP$_3$ & AP$_6$ & AP$_{12}$ & mAP &
AP$_3$ & AP$_6$ & AP$_{12}$ & AP$_{24}$ & mAP \\
\hline
\multirow{3}{*}{ViT-L}
 & CenterNet~\cite{CenterNet} & 11.74 & 26.45 & 45.92 & 61.64 & 36.43 & 15.87 & 37.24 & 62.63 & 78.80 & 48.63 & 11.63 & 31.13 & 59.90 & 77.22 & 44.97 \\
& MonoDGP~\cite{MonoDGP} &2.45  & 9.24  & 25.17 & 45.84 & 20.67 & 2.49  & 9.62  & 26.38 & 53.18 & 22.92 & 2.35  & 5.69  & 16.33 & 43.12 & 16.87 \\
& MonoLAA (Ours) & 16.60 & 30.92 & 45.92 & 60.84 & 38.57 & 28.30 & 48.23 & 65.80 & 77.90 & 55.06 & 19.06 & 32.72 & 53.53 & 71.21 & 44.13 \\
\hline
\multirow{3}{*}{ResNet152}
 & CenterNet~\cite{CenterNet} & 22.31 & 43.68 & 62.50 & 74.97 & 50.86 & 25.94 & 55.38 & 77.02 & 89.30 & 61.91 & 20.66 & 48.25 & 76.27 & 86.82 & 58.00 \\
& MonoDGP~\cite{MonoDGP} & 11.79 & 31.54 & 55.82 & 75.46 & 43.65 & 10.40 & 30.93 & 62.08 & 85.89 & 47.33 & 6.49 & 14.82 & 44.46 & 71.51 & 34.32 \\
& MonoLAA (Ours) & 43.38 & 54.37 & 64.86 & 73.90 & 59.13 & 55.66 & 72.29 & 82.47 & 89.76 & 75.05 & 58.12 & 66.61 & 72.31 & 83.41 & 70.11 \\
\hline
\multirow{3}{*}{ConvNeXt-L}
 & CenterNet~\cite{CenterNet} & 18.37 & 41.31 & 63.72 & 78.64 & 50.51 & 19.94 & 49.74 & 77.35 & 89.76 & 59.20 & 10.68 & 35.69 & 67.52 & 85.23 & 49.78 \\
& MonoDGP~\cite{MonoDGP} & 21.31 & 47.28 & 70.05 & 83.48 & 55.53 & 16.21 & 48.45 & 77.56 & 90.29 & 58.12 & 11.11 & 28.36 & 59.01 & 82.55 & 45.26 \\
& MonoLAA (Ours) & 27.96 & 48.41 & 65.89 & 77.97 & 55.06 & 32.29 & 58.89 & 80.18 & 89.25 & 65.15 & 21.76 & 41.96 & 71.97 & 86.23 & 55.48 \\
\hline
\end{tabular}
}
\caption{ Detailed 3D AP (\%) results on LAA3D-sim test set.}
\label{full_sim_test_ap}
\end{table*}

\begin{figure*}
  \centering
  \includegraphics[width=0.99\textwidth]{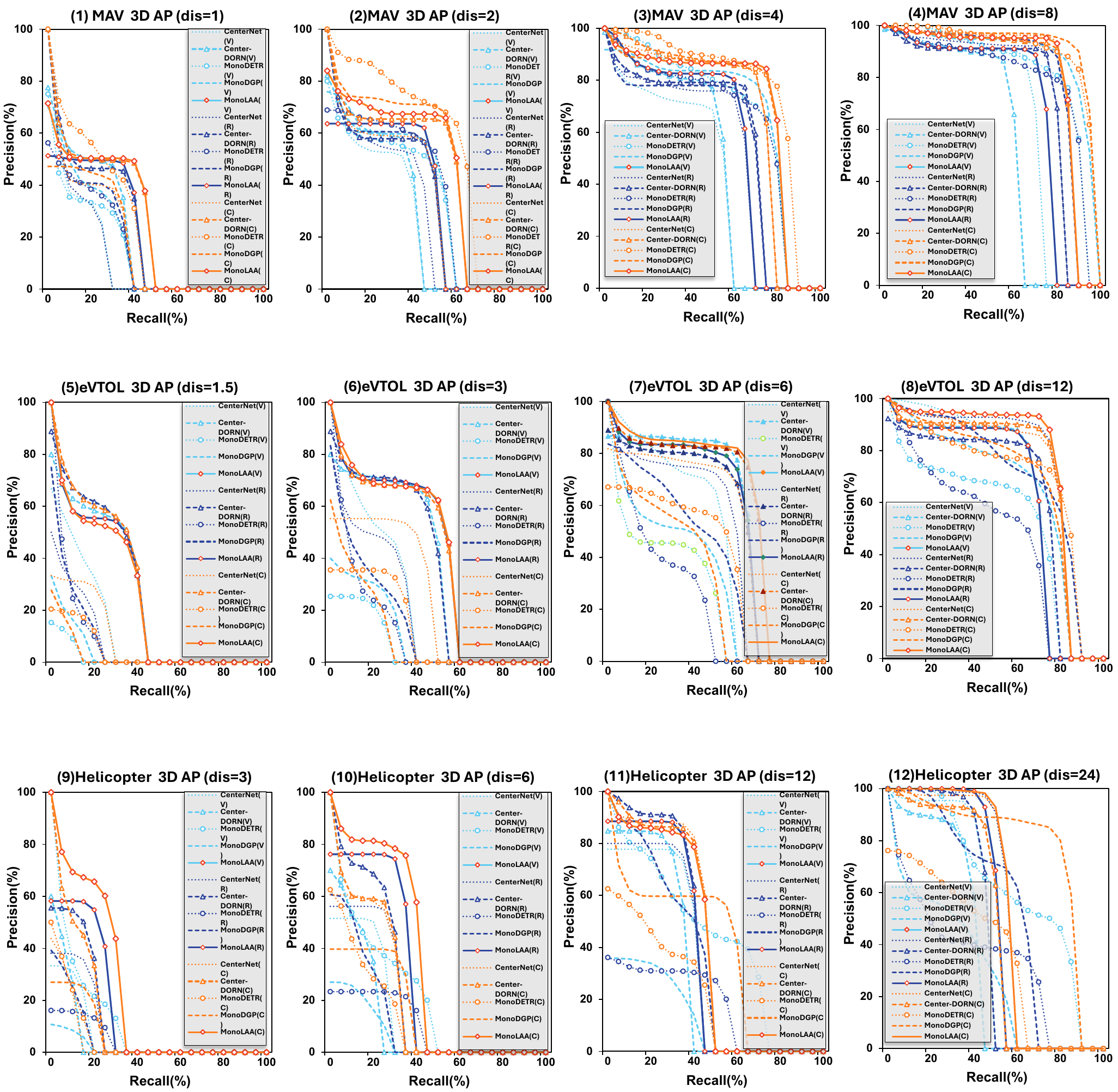}
  \caption{3D AP curve on LAA3D-real test set, where (V) denotes ViT-L, (R) denotes ResNet152 and (C) denotes ConvNeXt-L.}
  \label{app_real_ap}
\end{figure*}

\begin{figure*}
  \centering
  \includegraphics[width=0.99\textwidth]{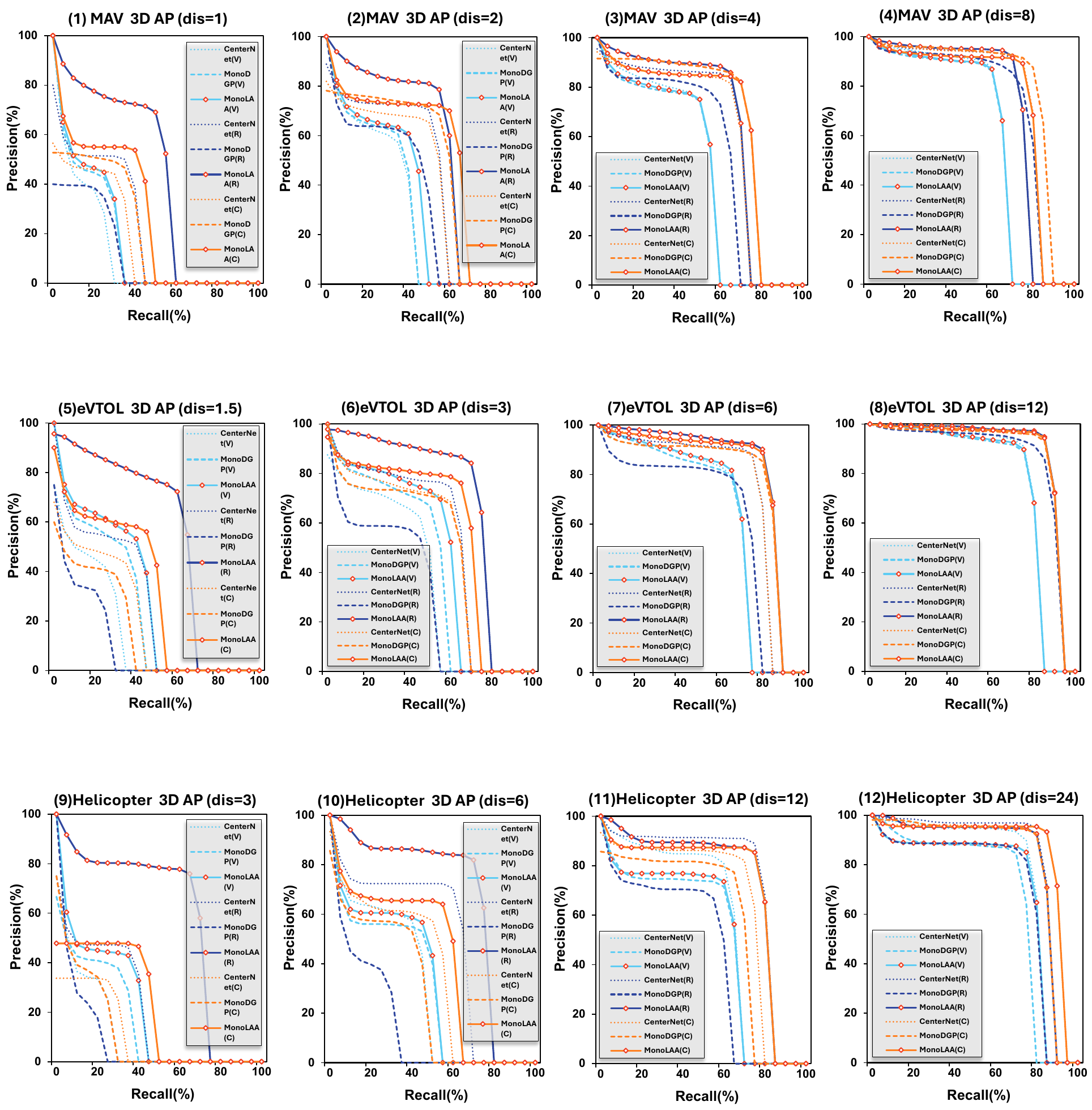}
  \caption{3D AP curve on LAA3D-sim test set, where (V) denotes ViT-L, (R) denotes ResNet152 and (C) denotes ConvNeXt-L.}
  \label{app_sim_ap}
\end{figure*}

\begin{table*}[htbp]
\centering
\small
\setlength{\tabcolsep}{2pt}
\resizebox{\textwidth}{!}{
\begin{tabular}{|l|l|ccccc|ccccc|ccccc|cccc|cc|}
\hline
\multirow{2}*{Backbone} & \multirow{2}*{Method} &
\multicolumn{5}{c|}{MAV} &
\multicolumn{5}{c|}{eVTOL} &
\multicolumn{5}{c|}{Helicopter} &
\multirow{2}*{mAOE↓} & \multirow{2}*{mATE↓} & \multirow{2}*{mASE↓} & \multirow{2}*{mDR↑} & \multirow{2}*{mAP↑} & \multirow{2}*{ADS↑} \\
\cline{3-17}
& & AOE↓ & ATE↓ & ASE↓ & DR↑ & 3DAP↑ &
AOE↓ & ATE↓ & ASE↓ & DR↑ & 3DAP↑ &
AOE↓ & ATE↓ & ASE↓ & DR↑ & 3DAP↑ &
 & & & & & \\
\hline
\multirow{5}{*}{ViT-L}
 & CenterNet~\cite{CenterNet} & 18.73 & 3.01  & 3.74  & 86.35 & 43.48 & 5.33  & 5.87  & 3.56  & 88.59 & 39.06 & 7.95  & 7.95  & 14.85 & 58.82 & 28.89 & 10.67 & 5.61  & 7.38  & 77.92 & 37.14 & 51.03 \\
         & Center-DORN~\cite{DORN} & 24.29 & 2.54  & 9.06  & 80.58 & 47.14 & 7.47  & 4.78  & 7.82  & 88.97 & 52.08 & 15.4  & 11.29 & 32.42 & 61.03 & 26.29 & 15.72 & 6.2   & 16.43 & 76.86 & 41.84 & 49.67 \\
         & MonoDETR~\cite{MonoDETR} & 25.48 & 2.28  & 2.22  & 96.78 & 50.69 & 9.69  & 8.66  & 6.44  & 92.38 & 25.26 & 22.35 & 13.34 & 14.36 & 94.23 & 24.27 & 19.17 & 8.09  & 7.67  & 94.46 & 33.41 & 48.3 \\
         & MonoDGP~\cite{MonoDGP} & 19.37 & 2.4   & 1.65  & 97.72 & 51.85 & 5.93  & 7.69  & 0.72  & 97.07 & 37.11 & 17.48 & 12.55 & 4.91  & 100   & 37.02 & 14.26 & 7.55  & 2.43  & 98.26 & 41.99 & 55.37 \\
         & MonoLAA(Ours) & 22.51 & 1.64  & 2.99  & 88.3  & 60.34 & 6.96  & 3.72  & 5.07  & 89.16 & 58.52 & 11.88 & 5.95  & 19.68 & 62.5  & 45.93 & 13.78 & 3.77  & 9.24  & 79.99 & 54.93 & 62.47 \\
    \hline
    \multirow{5}{*}{ResNet152}
    & CenterNet~\cite{CenterNet} & 13.75 & 2.21  & 3.33  & 95.16 & 53.49 & 6.06  & 5.4   & 4.4   & 93.35 & 39.44 & 9.16  & 8.33  & 8.47  & 72.06 & 32.34 & 9.66  & 5.31  & 5.4   & 86.86 & 41.75 & 56.25 \\
        & Center-DORN~\cite{DORN} & 20.27 & 2.12  & 10.94 & 92.59 & 55.71 & 8.7   & 5.4   & 10.42 & 89.73 & 53.9  & 13.28 & 7.96  & 21.41 & 63.24 & 34.85 & 14.08 & 5.16  & 14.26 & 81.85 & 48.15 & 55.61 \\
         & MonoDETR~\cite{MonoDETR} & 19.05 & 2.26  & 2.37  & 97.22 & 49.87 & 7.8   & 10.1  & 9.86  & 94.89 & 23.23 & 16.61 & 18.81 & 23.77 & 92.45 & 21.3  & 14.49 & 10.39 & 12    & 94.85 & 31.47 & 47.69 \\
        & MonoDGP~\cite{MonoDGP} & 19.37 & 2.4   & 1.65  & 97.72 & 51.85 & 5.93  & 7.69  & 0.72  & 97.07 & 37.11 & 17.48 & 12.55 & 4.91  & 93.68 & 37.02 & 14.26 & 7.55  & 2.43  & 96.15 & 41.99 & 55.37 \\
         & MonoLAA(Ours) & 18.72 & 1.6   & 1.42  & 95.32 & 64.59 & 8.19  & 4.43  & 4.96  & 93.35 & 53.69 & 9.83  & 6.44  & 13.3  & 70.59 & 46.24 & 12.24 & 4.16  & 6.56  & 86.42 & 54.84 & 63.7 \\
    \hline
    \multirow{5}{*}{ConvNeXt-L}
     & CenterNet~\cite{CenterNet} & 14.37 & 2.12  & 3.22  & 95.79 & 56.03 & 6.56  & 6.01  & 3.69  & 94.68 & 43.31 & 8.49  & 8.61  & 7.72  & 74.26 & 36.78 & 9.81  & 5.58  & 4.88  & 88.24 & 45.37 & 57.91 \\
        & Center-DORN~\cite{DORN} & 19.57 & 1.7   & 4.93  & 93.68 & 62.56 & 6.81  & 4.32  & 5.78  & 91.83 & 55.46 & 8.34  & 8.28  & 15.51 & 66.18 & 35.15 & 11.57 & 4.77  & 8.74  & 83.89 & 51.06 & 60.47 \\
          & MonoDETR~\cite{MonoDETR} & 18.46 & 1.83  & 1.56  & 98.74 & 59.79 & 5.89  & 5.89  & 5.84  & 93.36 & 34.04 & 17.89 & 16.84 & 17.42 & 96.54 & 23.85 & 14.08 & 8.18  & 8.27  & 96.21 & 39.23 & 53.15 \\
         & MonoDGP~\cite{MonoDGP} & 15.52 & 1.72  & 0.96  & 97.56 & 61.04 & 6.46  & 7.57  & 1.08  & 95.9  & 29.39 & 15.96 & 9     & 5.06  & 95.32 & 41.91 & 12.65 & 6.1   & 2.37  & 96.26 & 44.11 & 58.6 \\
         & MonoLAA(Ours) & 16.7  & 1.36  & 1.22  & 96.41 & 71.74 & 5.9   & 2.99  & 3.44  & 94.11 & 62.31 & 7.4   & 4.34  & 9.27  & 65.44 & 44.44 & 10    & 2.9   & 4.64  & 85.32 & 59.5  & 68.97 \\
    \hline
    \end{tabular}%
    }
        \caption{Detailed 3D object detection results on LAA3D-real validation set.}
  \label{full_real_val_results}%
\end{table*}%

\begin{table*}[htbp]
\centering
\small
\setlength{\tabcolsep}{6pt}
\resizebox{\textwidth}{!}{
\begin{tabular}{|l|l|ccccc|ccccc|ccccc|}
\hline
\multirow{2}*{Backbone} & \multirow{2}*{Method} &
\multicolumn{5}{c|}{MAV (3D AP↑)} &
\multicolumn{5}{c|}{eVTOL (3D AP↑)} &
\multicolumn{5}{c|}{Helicopter (3D AP↑)} \\
\cline{3-17}
 & & AP$_1$ & AP$_2$ & AP$_4$ & AP$_8$ & mAP &
 AP$_{1.5}$ & AP$_3$ & AP$_6$ & AP$_{12}$ & mAP &
 AP$_3$ & AP$_6$ & AP$_{12}$ & AP$_{24}$ & mAP \\
\hline
\multirow{5}{*}{ViT-L}
 & CenterNet~\cite{CenterNet} & 15.17 & 32.62 & 52.95 & 73.16 & 43.48 & 11.73 & 24.82 & 48.78 & 70.90 & 39.06 & 4.74 & 17.03 & 38.50 & 55.30 & 28.89 \\
 & Center-DORN~\cite{DORN} & 22.84 & 37.05 & 58.32 & 70.33 & 47.14 & 27.65 & 44.42 & 64.57 & 71.68 & 52.08 & 9.65 & 13.52 & 35.74 & 46.27 & 26.29 \\
 & MonoDETR~\cite{MonoDETR} & 10.84 & 32.89 & 70.90 & 88.15 & 50.69 & 3.27 & 10.35 & 30.37 & 57.05 & 25.26 & 3.28 & 7.45 & 32.49 & 53.84 & 24.27 \\
 & MonoDGP~\cite{MonoDGP} & 13.67 & 37.69 & 68.60 & 87.45 & 51.85 & 9.96 & 20.15 & 45.41 & 72.92 & 37.11 & 2.29 & 13.91 & 46.89 & 84.99 & 37.02 \\
 & MonoLAA(Ours) & 29.15 & 51.94 & 75.85 & 84.40 & 60.34 & 32.79 & 50.95 & 70.95 & 79.41 & 58.52 & 25.40 & 45.08 & 54.18 & 59.04 & 45.93 \\
\hline
\multirow{5}{*}{ResNet152}
 & CenterNet~\cite{CenterNet} & 13.18 & 34.63 & 74.59 & 91.55 & 53.49 & 8.79 & 21.10 & 49.27 & 78.60 & 39.44 & 7.12 & 16.76 & 34.37 & 71.10 & 32.34 \\
 & Center-DORN~\cite{DORN} & 23.52 & 44.16 & 73.55 & 81.62 & 55.71 & 33.25 & 47.54 & 61.62 & 73.17 & 53.90 & 12.98 & 25.97 & 45.51 & 54.96 & 34.85 \\
 & MonoDETR~\cite{MonoDETR} & 11.41 & 39.93 & 65.18 & 82.97 & 49.87 & 4.07 & 10.43 & 27.36 & 51.05 & 23.23 & 4.53 & 10.83 & 26.44 & 43.38 & 21.30 \\
 & MonoDGP~\cite{MonoDGP} & 13.67 & 37.69 & 68.60 & 87.45 & 51.85 & 9.96 & 20.15 & 45.41 & 72.92 & 37.11 & 2.29 & 13.91 & 46.89 & 84.99 & 37.02 \\
 & MonoLAA(Ours) & 31.69 & 57.69 & 80.07 & 88.89 & 64.59 & 31.86 & 44.31 & 63.13 & 75.46 & 53.69 & 20.25 & 40.62 & 58.15 & 65.93 & 46.24 \\
\hline
\multirow{5}{*}{ConvNeXt-L}
 & CenterNet~\cite{CenterNet} & 14.82 & 40.74 & 76.92 & 91.64 & 56.03 & 12.67 & 32.59 & 53.19 & 74.77 & 43.31 & 10.58 & 25.30 & 45.25 & 66.02 & 36.78 \\
 & Center-DORN~\cite{DORN} & 29.50 & 53.74 & 79.64 & 87.36 & 62.56 & 32.31 & 47.21 & 65.27 & 77.04 & 55.46 & 15.50 & 27.22 & 43.63 & 54.26 & 35.15 \\
 & MonoDETR~\cite{MonoDETR} & 17.71 & 49.37 & 81.13 & 90.94 & 59.79 & 5.06 & 16.56 & 43.31 & 71.24 & 34.04 & 4.95 & 15.61 & 23.35 & 51.49 & 23.85 \\
 & MonoDGP~\cite{MonoDGP} & 17.33 & 50.41 & 83.64 & 92.79 & 61.04 & 3.59 & 11.91 & 31.81 & 70.24 & 29.39 & 6.18 & 21.40 & 50.11 & 89.96 & 41.91 \\
 & MonoLAA(Ours) & 38.12 & 69.60 & 86.22 & 93.03 & 71.74 & 28.88 & 55.67 & 77.30 & 87.39 & 62.31 & 18.17 & 32.18 & 62.10 & 65.29 & 44.44 \\
\hline
\end{tabular}}
\caption{Detailed 3D AP results on LAA3D-real validation set.}
  \label{full_real_val_ap}%
\end{table*}

\begin{table*}[htbp]
\centering
\small
\setlength{\tabcolsep}{2pt}
\resizebox{\textwidth}{!}{
\begin{tabular}{|l|l|ccccc|ccccc|ccccc|cccc|cc|}
\hline
\multirow{2}*{Backbone} & \multirow{2}*{Method} &
\multicolumn{5}{c|}{MAV} &
\multicolumn{5}{c|}{eVTOL} &
\multicolumn{5}{c|}{Helicopter} &
\multirow{2}*{mAOE↓} & \multirow{2}*{mATE↓} & \multirow{2}*{mASE↓} & \multirow{2}*{mDR↑} & \multirow{2}*{mAP↑} & \multirow{2}*{ADS↑} \\
\cline{3-17}
& & AOE↓ & ATE↓ & ASE↓ & DR↑ & 3DAP↑ &
AOE↓ & ATE↓ & ASE↓ & DR↑ & 3DAP↑ &
AOE↓ & ATE↓ & ASE↓ & DR↑ & 3DAP↑ &
 & & & & & \\
\hline
\multirow{3}*{ViT-L} 
& CenterNet~\cite{CenterNet} & 26.72 & 3.47 & 5.66 & 69.75 & 35.28 & 12.17 & 5.73 & 10.05 & 86.48 & 44.88 & 6.36 & 12.04 & 45.02 & 91.53 & 45.86 & 15.08 & 7.08 & 20.24 & 82.58 & 42.01 & 47.81 \\
& MonoDGP~\cite{MonoDGP} & 30.73 & 4.84  & 6.60   & 77.26 & 18.83 & 14.86 & 8.14  & 10.67 & 93.97 & 23.45 & 13.02 & 27.6  & 66.43 & 88.12 & 15.67 & 19.53 & 13.53 & 27.9  & 86.45 & 19.32 & 34.43 \\
& MonoLAA (Ours) & 27.39 & 3.52 & 5.99 & 69.52 & 37.31 & 12.55 & 5.40 & 11.01 & 85.86 & 49.84 & 8.17 & 13.26 & 49.15 & 90.17 & 42.22 & 16.04 & 7.40 & 22.05 & 81.85 & 43.12 & 47.74 \\
\hline
\multirow{3}*{ResNet152} 
& CenterNet~\cite{CenterNet} & 18.66 & 2.12 & 1.97 & 82.68 & 51.80 & 5.99 & 3.33 & 4.93 & 92.31 & 60.38 & 4.45 & 7.01 & 13.27 & 94.24 & 57.69 & 9.70 & 4.15 & 6.72 & 89.74 & 56.62 & 65.70 \\
& MonoDGP~\cite{MonoDGP} & 20.62 & 3.04 & 2.80 & 85.17 & 44.44 & 8.34 & 4.94 & 3.41 & 97.09 & 46.69 & 7.32 & 15.07 & 24.10 & 93.98 & 33.00 & 12.10 & 7.68 & 10.10 & 92.08 & 41.38 & 53.06 \\
& MonoLAA (Ours) & 18.56 & 1.72 & 2.55 & 79.54 & 59.21 & 6.76 & 2.77 & 3.73 & 92.31 & 69.38 & 5.20 & 5.65 & 15.18 & 92.54 & 67.10 & 10.17 & 3.38 & 7.15 & 88.13 & 65.23 & 70.84 \\
\hline
\multirow{3}*{ConvNeXt-L} 
& CenterNet~\cite{CenterNet} & 17.72 & 2.20 & 1.98 & 87.23 & 52.02 & 7.29 & 4.03 & 5.85 & 94.42 & 55.18 & 5.44 & 7.63 & 12.24 & 94.58 & 52.34 & 10.15 & 4.62 & 6.69 & 92.07 & 53.38 & 63.47 \\
& MonoDGP~\cite{MonoDGP} & 16.17 & 2.08 & 1.27 & 90.40 & 57.12 & 5.61 & 3.36 & 1.54 & 97.56 & 55.51 & 5.59 & 7.67 & 1.89 & 95.11 & 47.12 & 9.12 & 4.37 & 1.57 & 94.36 & 53.25 & 65.83 \\
& MonoLAA (Ours) & 19.55 & 1.98 & 3.19 & 82.05 & 58.42 & 8.59 & 3.24 & 7.50 & 93.18 & 67.04 & 6.52 & 6.21 & 18.38 & 91.86 & 63.35 & 11.56 & 3.81 & 9.69 & 89.03 & 62.94 & 67.99 \\
\hline
\end{tabular}}
        \caption{Detailed 3D object detection results on LAA3D-sim validation set.}
\label{full_sim_val_results}
\end{table*}

\begin{table*}[htbp]
\centering
\small
\setlength{\tabcolsep}{6pt}
\renewcommand{\arraystretch}{1.1}
\resizebox{\textwidth}{!}{
\begin{tabular}{|l|l|ccccc|ccccc|ccccc|}
\hline
\multirow{2}*{Backbone} & \multirow{2}*{Method} &
\multicolumn{5}{c|}{MAV 3D AP↑} &
\multicolumn{5}{c|}{eVTOL 3D AP↑} &
\multicolumn{5}{c|}{Helicopter 3D AP↑} \\
\cline{3-17}
& & AP$_1$ & AP$_2$ & AP$_4$ & AP$_5$ & mAP &
AP$_{1.5}$ & AP$_3$ & AP$_6$ & AP$_{12}$ & mAP &
AP$_3$ & AP$_6$ & AP$_{12}$ & AP$_{24}$ & mAP \\
\hline
\multirow{3}{*}{ViT-L}
 & CenterNet~\cite{CenterNet} & 10.74 & 24.02 & 45.18 & 61.20 & 35.28 & 13.51 & 32.80 & 58.45 & 74.76 & 44.88 & 12.12 & 30.48 & 63.50 & 77.33 & 45.86 \\
& MonoDGP~\cite{MonoDGP} & 2.64  & 8.71  & 22.02 & 41.93 & 18.83 & 2.71  & 9.89  & 28.14 & 53.09 & 23.45 & 1.23  & 3.82  & 13.84 & 43.78 & 15.67 \\
& MonoLAA (Ours) & 13.62 & 29.19 & 45.84 & 60.59 & 37.31 & 23.42 & 43.09 & 60.26 & 72.59 & 49.84 & 15.60 & 30.59 & 51.68 & 71.00 & 42.22 \\
\hline
\multirow{3}{*}{ResNet152}
 & CenterNet~\cite{CenterNet} & 23.25 & 44.25 & 63.62 & 76.09 & 51.80 & 25.95 & 51.50 & 76.64 & 87.41 & 60.38 & 18.41 & 46.79 & 75.11 & 90.44 & 57.69 \\
& MonoDGP~\cite{MonoDGP} & 12.41 & 31.40 & 57.35 & 76.58 & 44.44 & 8.29 & 30.71 & 62.70 & 85.08 & 46.69 & 3.66 & 17.15 & 39.63 & 71.58 & 33.00 \\
& MonoLAA (Ours) & 42.65 & 55.47 & 64.79 & 73.92 & 59.21 & 47.75 & 65.65 & 78.28 & 85.85 & 69.38 & 52.91 & 64.97 & 71.32 & 79.19 & 67.10 \\
\hline
\multirow{3}{*}{ConvNeXt-L}
 & CenterNet~\cite{CenterNet} & 19.30 & 42.80 & 66.55 & 79.42 & 52.02 & 17.45 & 46.22 & 72.75 & 86.65 & 55.77 & 14.15 & 36.21 & 70.84 & 88.17 & 52.34 \\
& MonoDGP~\cite{MonoDGP} & 21.93 & 48.07 & 72.95 & 85.53 & 57.12 & 14.54 & 40.86 & 75.50 & 91.14 & 55.51 & 9.47 & 28.31 & 62.47 & 88.23 & 47.12 \\
& MonoLAA (Ours) & 39.95 & 53.80 & 64.64 & 75.31 & 58.42 & 43.16 & 62.41 & 76.79 & 85.82 & 67.04 & 44.05 & 59.08 & 70.29 & 79.98 & 63.35 \\
\hline
\end{tabular}
}
\caption{Detailed 3D AP results on LAA3D-sim validation set.}
\label{full_sim_val_ap}
\end{table*}

\begin{figure*}
  \centering
  \includegraphics[width=0.99\textwidth]{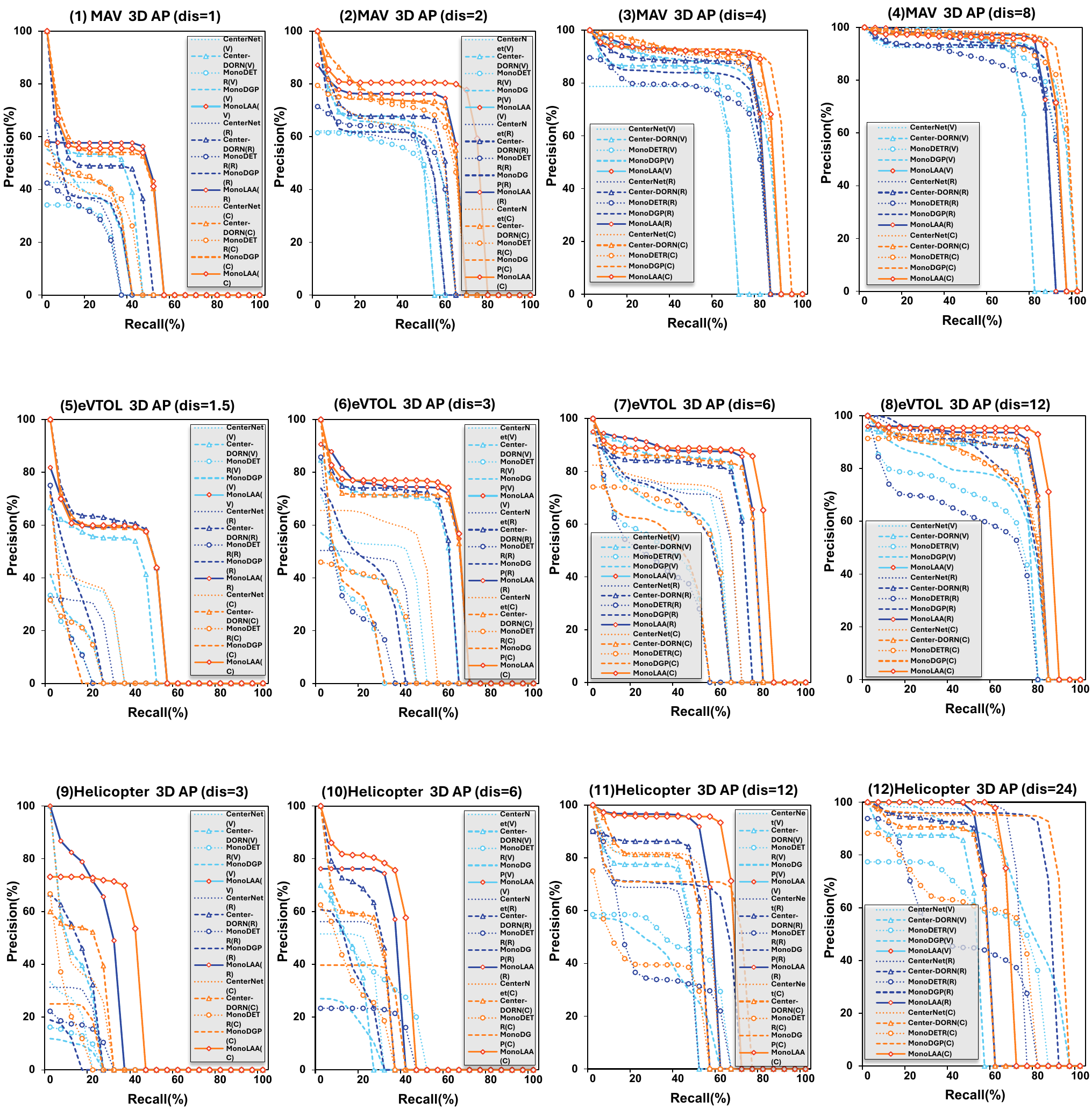}
  \caption{3D AP curve on LAA3D-real validation set, where (V) denotes ViT-L, (R) denotes ResNet152 and (C) denotes ConvNeXt-L.}
  \label{app_real_ap_val}
\end{figure*}

\begin{figure*}
  \centering
  \includegraphics[width=0.99\textwidth]{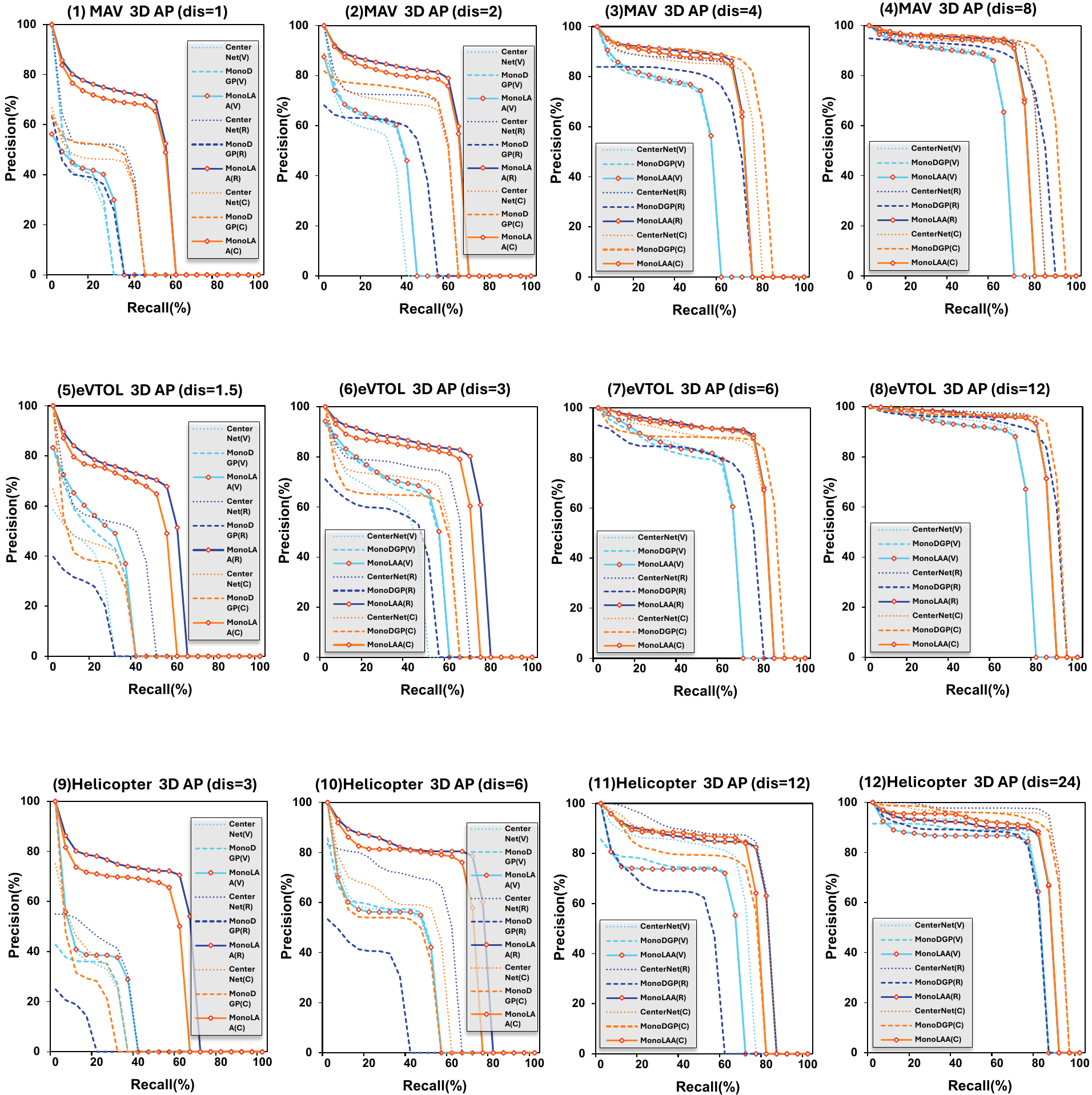}
  \caption{3D AP curve on LAA3D-sim validation set, where (V) denotes ViT-L, (R) denotes ResNet152 and (C) denotes ConvNeXt-L.}
  \label{app_sim_ap_val}
\end{figure*}

\begin{table*}[htbp]
\centering
\small
\setlength{\tabcolsep}{2pt}
\renewcommand{\arraystretch}{1.15}
\resizebox{\textwidth}{!}{
\begin{tabular}{|l|l|
c c c c c|
c c c c c|
c c c c c|
c c c c |c c|}
\hline
\multirow{2}*{Setting} & \multirow{2}*{Method} &
\multicolumn{5}{c|}{MAV} &
\multicolumn{5}{c|}{eVTOL} &
\multicolumn{5}{c|}{Helicopter} &
\multirow{2}*{mAOE↓} & \multirow{2}*{mATE↓} & \multirow{2}*{mASE↓} & \multirow{2}*{mDR↑} & \multirow{2}*{mAP↑} & \multirow{2}*{ADS↑}\\
\cline{3-17}
& & AOE↓ & ATE↓ & ASE↓ & DR↑ & 3DAP↑ 
& AOE↓ & ATE↓ & ASE↓ & DR↑ & 3DAP↑
& AOE↓ & ATE↓ & ASE↓ & DR↑ & 3DAP↑
& & & & & & \\

\hline
\multirow{1}{*}{Supervised} 
& MonoLAA
& 22.02 & 2.52 & 1.83 & 92.08 & 49.43 
& 7.80 & 5.16 & 4.98 & 91.41 & 47.53 
& 10.26 & 6.37 & 17.36 & 56.52 & 34.34 
& 13.36 & 4.69 & 8.05 & 80.00 & 43.77 & 55.23 \\
\hline
\multirow{3}{*}{UDA}
& Direct Test 
& 40.87 & 10.91 & 42.90 & 60.62 & 12.37 
& 40.69 & 20.84 & 144.29 & 42.75 & 2.05 
& 40.04 & 54.84 & 257.47 & 2.17 & 0.06 
& 40.53 & 28.87 & 148.22 & 35.18 & 4.83 & 8.64 \\
& BG Replace~\cite{MAV6D} 
& 44.19 & 10.93 & 30.33 & 45.77 & 10.24 
& 42.93 & 22.19 & 138.09 & 43.32 & 1.72 
& 22.12 & 46.77 & 409.16 & 3.62 & 0.02 
& 36.41 & 26.63 & 192.53 & 30.91 & 4.00 & 9.89 \\
& Pseudo-label~\cite{ST3D} 
& 42.58 & 11.13 & 43.94 & 78.29 & 14.70 
& 37.72 & 18.76 & 140.33 & 60.11 & 4.57 
& 38.10 & 60.70 & 134.31 & 10.14 & 0.00 
& 39.47 & 30.20 & 106.19 & 49.52 & 6.42 & 11.44 \\
\hline
\multirow{1}{*}{DA}
& Fine-tuning  &
17.25 & 1.67 & 4.66 & 95.20 & 68.43 
& 5.78 & 3.37 & 6.55 & 93.51 & 57.51 
& 11.16 & 4.79 & 17.85 & 54.35 & 35.70 
& 11.40 & 3.28 & 9.69 & 81.02 & 53.88 & 63.24 \\
\hline
\end{tabular}}
\caption{Detailed domain adaptation results on the LAA3D-real test set.
}
\label{full_test_da}
\end{table*}

\begin{table*}[htbp]
\centering
\small
\setlength{\tabcolsep}{2pt}
\renewcommand{\arraystretch}{1.15}
\resizebox{\textwidth}{!}{
\begin{tabular}{|l|l|
c c c c c|
c c c c c|
c c c c c|
c c c c |c c|}
\hline
\multirow{2}*{Setting} & \multirow{2}*{Method} &
\multicolumn{5}{c|}{MAV} &
\multicolumn{5}{c|}{eVTOL} &
\multicolumn{5}{c|}{Helicopter} &
\multirow{2}*{mAOE↓} & \multirow{2}*{mATE↓} & \multirow{2}*{mASE↓} & \multirow{2}*{mDR↑} & \multirow{2}*{mAP↑} & \multirow{2}*{ADS↑}\\
\cline{3-17}
& & AOE↓ & ATE↓ & ASE↓ & DR↑ & 3DAP↑ 
& AOE↓ & ATE↓ & ASE↓ & DR↑ & 3DAP↑
& AOE↓ & ATE↓ & ASE↓ & DR↑ & 3DAP↑
& & & & & & \\

\hline
Supervised & MonoLAA &
18.72 & 1.60 & 1.42 & 95.32 & 64.59 &
8.19 & 4.43 & 4.96 & 93.35 & 53.69 &
9.83 & 6.44 & 13.30 & 70.59 & 46.24 &
12.24 & 4.16 & 6.56 & 86.42 & 54.84 & 63.70 \\
\hline
\multirow{3}*{UDA} & Direct test &
41.74 & 10.07 & 36.63 & 69.66 & 14.36 &
40.58 & 19.01 & 143.40 & 42.40 & 1.88 &
36.27 & 62.88 & 418.17 & 5.88 & 0.00 &
39.53 & 30.65 & 199.40 & 39.31 & 5.41 & 10.25 \\
 & BG replace~\cite{MAV6D} &
42.82 & 10.82 & 29.27 & 54.91 & 8.78 &
46.00 & 20.91 & 141.09 & 45.25 & 2.37 &
32.36 & 25.54 & 290.78 & 10.29 & 0.01 &
40.39 & 19.09 & 153.71 & 36.82 & 3.72 & 9.56 \\
 & Pseudo-label~\cite{ST3D} &
43.33 & 10.48 & 38.78 & 81.28 & 15.72 &
38.18 & 18.46 & 142.92 & 61.03 & 4.94 &
27.68 & 56.01 & 157.62 & 11.03 & 0.00 &
36.40 & 28.32 & 113.10 & 51.11 & 6.89 & 13.16 \\
\hline
Fine tuning &  &
13.63 & 1.16 & 3.06 & 97.19 & 76.00 &
6.27 & 2.61 & 6.26 & 95.44 & 63.95 &
11.44 & 5.50 & 21.09 & 64.71 & 46.82 &
10.45 & 3.09 & 10.14 & 85.78 & 62.25 & 68.98 \\
\hline
\end{tabular}
}
\caption{Domain adaptation results on the LAA3D-real validation set.}
\label{full_val_da}
\end{table*}

\begin{table*}[htbp]
\centering
\footnotesize
\setlength{\tabcolsep}{4pt}
\begin{tabular}{|l|l|l|cc|cc|cc|}
\hline
\multirow{2}*{Method} & \multirow{2}*{Backbone} & \multirow{2}*{Type} &
\multicolumn{2}{c|}{MAV} &
\multicolumn{2}{c|}{eVTOL} &
\multicolumn{2}{c|}{Helicopter} \\
\cline{4-9}
 &  &  & ADD↑ & ADD-S↑ & ADD↑ & ADD-S↑ & ADD↑ & ADD-S↑ \\
\hline
Yolo-6D~\cite{YOLO6D} & Darknet-19 & PnP & 21.51 & 32.99 & 60.54 & 73.15 & 61.62 & 75.58 \\
YOLOX-6D-Pose~\cite{Yolo-6d-pose} & CSPDarknet-53 & Direct & 24.42 & 33.73 & 56.20 & 69.93 & 57.40 & 75.00 \\
YOLOV5-6D~\cite{YOLOV5-6D} & CSPDarknet-53 & PnP & 29.74 & 38.21 & 66.10 & 76.69 & 63.85 & 80.72 \\
\hline
\end{tabular}
\caption{6D pose estimation results on the LAA3D-real validation set using ADD and ADD-S metrics.}
\label{pose_val}
\end{table*}

\begin{table*}[htbp]
\centering
\small
\setlength{\tabcolsep}{3pt}
\renewcommand{\arraystretch}{1.1}
\resizebox{\textwidth}{!}{
\begin{tabular}{|l|l|cccc|cccc|ccc|c|}
\hline
\multirow{2}*{Tracker} & \multirow{2}*{Detector (ResNet152)} &
\multicolumn{4}{c|}{CLEAR MOT Metrics} &
\multicolumn{4}{c|}{Identity Metrics} &
\multicolumn{3}{c|}{HOTA Metrics} &
\multirow{2}*{Frag} \\
\cline{3-13}
 & & MOTA↑ & MOTP↓ & MODA↑ & IDSWs↓ & IDF1↑ & IDTP↑ & IDFP↓ & IDFN↓ & HOTA↑ & DetA↑ & AssA↑ & \\
\hline
\multicolumn{14}{|c|}{MAV} \\
\hline
\multirow{5}*{AB3DMOT~\cite{AB3DMOT} } 
& CenterNet~\cite{CenterNet} &    37.62 & 1.47  & 39.44 & 50    & 66.44 & 1646  & 563   & 1100  & 47.67 & 47.18 & 48.46 & 34 \\
 & Center-DORN~\cite{DORN} & 49.93 & 1.34  & 51.75 & 50    & 74.33 & 1918  & 497   & 828   & 54.03 & 53.63 & 54.68 & 37 \\
 & MonoDETR~\cite{MonoDETR} & 38.16 & 1.29  & 40.42 & 62    & 63.81 & 1442  & 332   & 1304  & 43.23 & 42.32 & 44.68 & 27 \\
 & MonoDGP~\cite{MonoDGP} & 40.82 & 1.29  & 42.17 & 37    & 67.84 & 1675  & 517   & 1071  & 53.8  & 53.23 & 54.8  & 26 \\
 & MonoLAA (Ours) & 57.03 & 1.2   & 59.1  & 57    & 78.42 & 2040  & 417   & 706   & 56.22 & 55.9  & 56.74 & 37 \\
\hline
\multicolumn{14}{|c|}{eVTOL} \\
\hline
\multirow{5}*{AB3DMOT~\cite{AB3DMOT} } 
& CenterNet~\cite{CenterNet} & 14.85 & 2.57  & 15.71 & 15    & 52.99 & 832   & 557   & 919   & 10.71 & 9.7   & 12.68 & 10 \\
 & Center-DORN~\cite{DORN} & 33.87 & 1.34  & 34.38 & 9     & 63.79 & 1012  & 410   & 739   & 11.04 & 10.17 & 12.6  & 7 \\
 & MonoDETR~\cite{MonoDETR} & -22.04 & 2.87  & -21.47 & 10    & 23.02 & 318   & 694   & 1433  & 7.75  & 6.88  & 9.76  & 5 \\
 & MonoDGP~\cite{MonoDGP} & -12.96 & 2.75 & -12.56 & 7 & 35.44 & 541 & 761 & 1210 & 11.85 & 10.87 & 14.10 & 2 \\
 & MonoLAA (Ours) & 35.24 & 1.31  & 36.09 & 15    & 64.08 & 998   & 366   & 753   & 11.39 & 10.44 & 12.99 & 9 \\
\hline
\multicolumn{14}{|c|}{Helicopter} \\
\hline
\multirow{5}*{AB3DMOT~\cite{AB3DMOT} } 
& CenterNet~\cite{CenterNet} & 7.27  & 4.36  & 8.36  & 3     & 24.1  & 40    & 17    & 235   & 1.01  & 1.01  & 1.01  & 3 \\
 & Center-DORN~\cite{DORN} & 13.09 & 4.21  & 14.18 & 3     & 28.05 & 46    & 7     & 229   & 1.16  & 1.16  & 1.16  & 3 \\
 & MonoDETR~\cite{MonoDETR} & -13.82 & 5.6   & -12   & 5     & 27.36 & 58    & 91    & 217   & 1.34  & 1.33  & 1.35  & 3 \\
 & MonoDGP~\cite{MonoDGP} & -7.64 & 4.77  & -4.73 & 8     & 47.06 & 128   & 141   & 147   & 3.1   & 3.09  & 3.1   & 2 \\
 & MonoLAA (Ours) & 35.64 & 3.26  & 36.73 & 3     & 59.15 & 126   & 25    & 149   & 3.46  & 3.46  & 3.46  & 2 \\
\hline
\end{tabular}
}
\caption{Detailed MOT results on the LAA3D-real test set.}
\label{mot_test}
\end{table*}

\begin{figure*}
  \centering
  \includegraphics[width=0.99\textwidth]{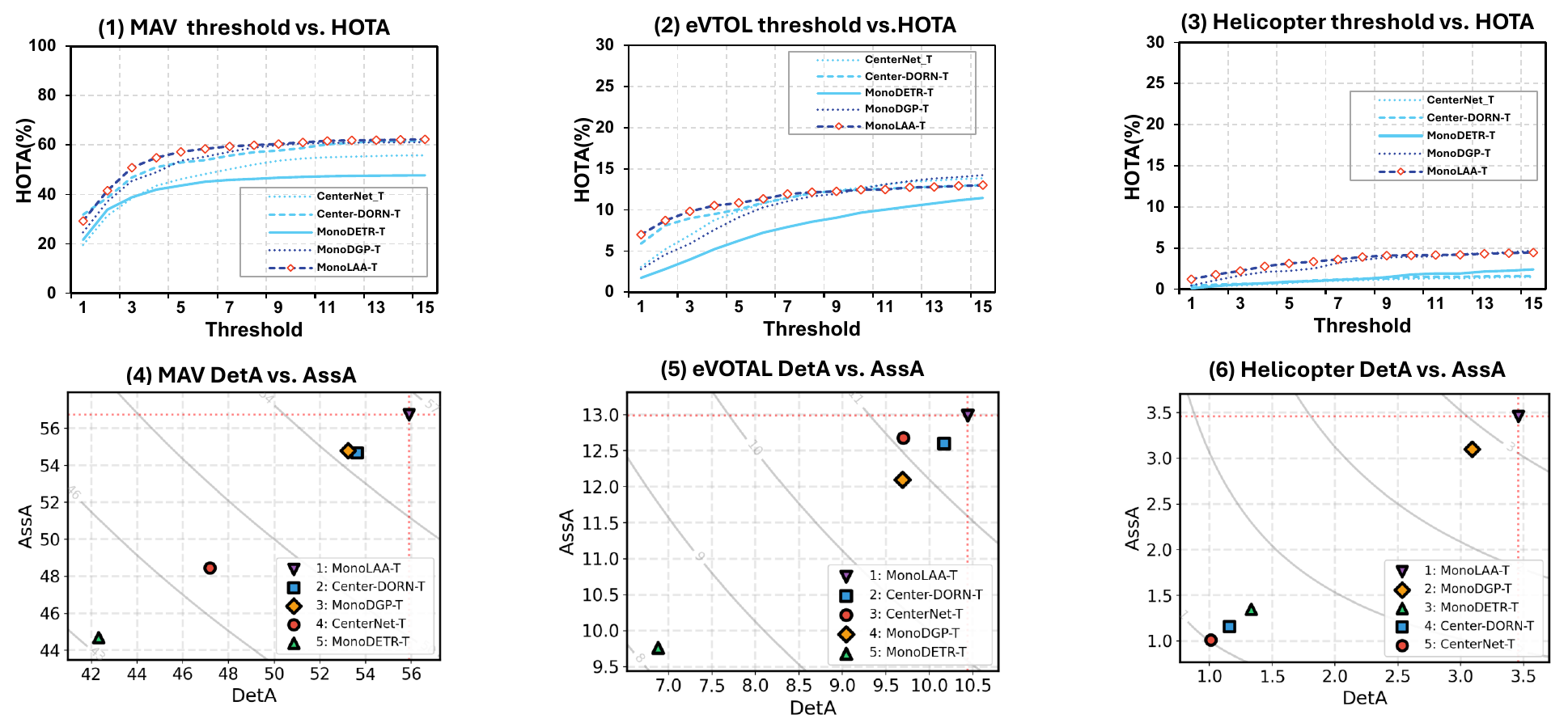}
  \caption{HOTA curve on LAA3D-real test set, where CenterNet-T, Center-DORN-T, MonoDETR-T, MonoDGP-T, and MonoLAA-T represent the MOT methods using AB3DMOT tracker and CenterNet,  Center-DORN, MonoDETR, MonoDGP and MonoLAA detectors, respectively.}
  \label{hota_test}
\end{figure*}

\begin{table*}[htbp]
\centering
\small
\setlength{\tabcolsep}{3pt}
\renewcommand{\arraystretch}{1.1}
\resizebox{\textwidth}{!}{
\begin{tabular}{|l|l|cccc|cccc|ccc|c|}
\hline
\multirow{2}{*}{Tracker} & \multirow{2}{*}{Detector (ResNet152)} &
\multicolumn{4}{c|}{CLEAR MOT Metrics} &
\multicolumn{4}{c|}{Identity Metrics} &
\multicolumn{3}{c|}{HOTA Metrics} &
\multirow{2}{*}{Frag} \\
\cline{3-13}
& & MOTA$\uparrow$ & MOTP$\downarrow$ & MODA$\uparrow$ & IDSWs$\downarrow$ & IDF1$\uparrow$ & IDTP$\uparrow$ & IDFP$\downarrow$ & IDFN$\downarrow$ & HOTA$\uparrow$ & DetA$\uparrow$ & AssA$\uparrow$ & \\
\hline
\multicolumn{14}{|c|}{MAV} \\
\hline
\multirow{5}*{AB3DMOT~\cite{AB3DMOT} }  & CenterNet~\cite{CenterNet} & 60.62 & 1.50 & 62.04 & 38 & 79.60 & 1980 & 321 & 694 & 55.35 & 54.85 & 56.14 & 28 \\
& Center-DORN~\cite{DORN} & 69.18 & 1.09 & 70.27 & 29 & 84.42 & 2154 & 275 & 520 & 57.43 & 57.13 & 57.87 & 22 \\
& MonoDETR~\cite{MonoDETR} & 48.50 & 1.50 & 50.82 & 62 & 71.86 & 1679 & 320 & 995 & 50.10 & 49.02 & 51.73 & 24 \\
& MonoDGP~\cite{MonoDGP} & 52.24 & 1.39 & 53.14 & 24 & 75.28 & 1908 & 487 & 766 & 59.31 & 58.70 & 60.29 & 18 \\
& MonoLAA & 74.94 & 0.92 & 76.29 & 36 & 87.76 & 2272 & 232 & 402 & 61.88 & 61.64 & 62.24 & 26 \\
\hline
\multicolumn{14}{|c|}{eVTOL} \\
\hline
\multirow{5}*{AB3DMOT~\cite{AB3DMOT} }  & CenterNet~\cite{CenterNet} & 21.18 & 2.46 & 22.39 & 21 & 56.71 & 881 & 493 & 852 & 11.50 & 10.39 & 13.50 & 14 \\
& Center-DORN~\cite{DORN} & 41.14 & 1.19 & 41.60 & 8 & 66.40 & 1000 & 279 & 733 & 11.77 & 10.83 & 13.35 & 6 \\
& MonoDETR~\cite{MonoDETR} & -18.12 & 2.67 & -17.48 & 11 & 24.31 & 327 & 630 & 1406 & 7.55 & 6.65 & 9.65 & 7 \\
& MonoDGP~\cite{MonoDGP} & 11.48 & 2.29 & 11.54 & 1 & 34.57 & 405 & 205 & 1328 & 10.91 & 10.10 & 12.64 & 1 \\
& MonoLAA & 41.60 & 1.18 & 41.89 & 5 & 67.10 & 1027 & 301 & 706 & 12.60 & 11.65 & 14.15 & 3 \\
\hline
\multicolumn{14}{|c|}{Helicopter} \\
\hline
\multirow{5}*{AB3DMOT~\cite{AB3DMOT} }  & CenterNet~\cite{CenterNet} & 1.84 & 5.34 & 4.41 & 7 & 25.29 & 44 & 32 & 228 & 1.07 & 1.07 & 1.07 & 7 \\
& Center-DORN~\cite{DORN} & 9.93 & 3.33 & 11.40 & 4 & 27.63 & 46 & 15 & 226 & 1.27 & 1.27 & 1.27 & 3 \\
& MonoDETR~\cite{MonoDETR} & -17.28 & 5.13 & -15.81 & 4 & 20.65 & 41 & 84 & 231 & 1.01 & 1.01 & 1.01 & 3 \\
& MonoDGP~\cite{MonoDGP} & 17.65 & 5.85 & 21.69 & 11 & 60.34 & 162 & 103 & 110 & 3.60 & 3.60 & 3.62 & 7 \\
& MonoLAA & 49.26 & 3.24 & 50.74 & 4 & 70.35 & 159 & 21 & 113 & 4.36 & 4.36 & 4.36 & 3 \\
\hline
\end{tabular}
}
\caption{Detailed MOT results on the LAA3D-real validation set.}
\label{mot_val}
\end{table*}

\begin{figure*}
  \centering
  \includegraphics[width=0.99\textwidth]{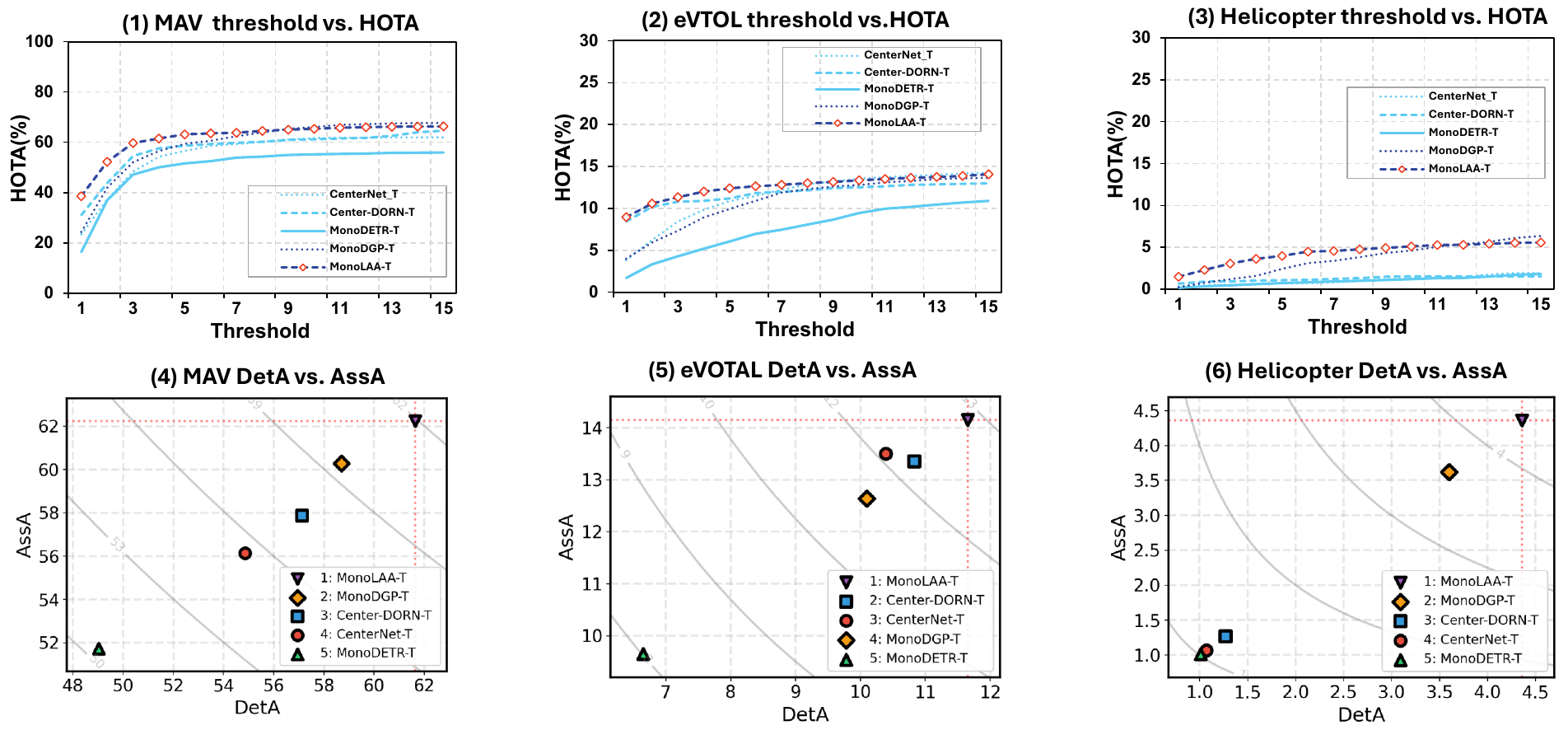}
  \caption{HOTA curve on LAA3D-real validation set, where CenterNet-T, Center-DORN-T, MonoDETR-T, MonoDGP-T, and MonoLAA-T represent the MOT methods using AB3DMOT tracker and CenterNet,  Center-DORN, MonoDETR, MonoDGP and MonoLAA detectors, respectively.}
  \label{hota_val}
\end{figure*}